\documentclass[sigconf,balance=false]{acmart}
\usepackage{popets}
\usepackage{soul}
\usepackage{makecell}

\usepackage{multirow,gensymb}
\usepackage{mathtools}
\usepackage{enumitem}

\usepackage{subcaption}

\newcommand\Mycomb[2][^n]{\prescript{#1\mkern-0.5mu}{}C_{#2}}

\usepackage{algorithm,amsmath} 
\usepackage{algpseudocode}
\usepackage{threeparttable,booktabs}

\usepackage[normalem]{ulem}

\setcopyright{popets}
\copyrightyear{2026}

\acmYear{2026}
\acmVolume{}
\acmNumber{}
\acmDOI{}
\acmISBN{}
\acmConference{Proceedings on Privacy Enhancing Technologies}
\begin{document}

\title[SilhouetteTell]{SilhouetteTell: Practical Video Identification ~\\Leveraging Blurred Recordings of Video Subtitles}

\author{Guanchong Huang}
\affiliation{%
  \institution{\textit{School of Computer Science}\\
               \textit{University of Oklahoma}}
  \city{Norman}
  \state{OK}
  \country{USA}
}
\email{guanchong.huang@ou.edu}

\author{Song Fang}
\affiliation{%
  \institution{\textit{School of Computer Science}\\
               \textit{University of Oklahoma}}
  \city{Norman}
  \state{OK}
  \country{USA}
}
\email{songf@ou.edu}

% \thanks{\textsuperscript{*}Corresponding author}

\begin{abstract}
Video identification attacks pose a significant privacy threat that can reveal videos that victims watch, which may disclose their hobbies, religious beliefs, political leanings, sexual orientation, and health status. Also, video watching history can be used for user profiling or advertising and may result in cyberbullying, discrimination, or blackmail. Existing extensive video inference techniques usually depend on analyzing network traffic generated by streaming online videos. In this work, we observe that the content of a subtitle determines its silhouette displayed on the screen, and identifying each subtitle silhouette also derives the temporal difference between two consecutive subtitles.  We then propose~\textit{SilhouetteTell}, a novel video identification attack that combines the spatial and time domain information into a spatiotemporal feature of subtitle silhouettes. \textit{SilhouetteTell} explores the spatiotemporal correlation between recorded subtitle silhouettes of a video and its subtitle file. It can infer both online and offline videos. Comprehensive experiments on off-the-shelf smartphones confirm the high efficacy of \textit{SilhouetteTell} for inferring video titles and clips under various settings, including from a distance of up to 40 meters. 
\end{abstract}

% Privacy, 

\maketitle

\section{Introduction}
\label{sec:introduction} 

Video viewing has become increasingly popular. According to a survey with 528 unique respondents conducted in November 2022, people on average watch 17 hours of videos per week~\cite{WyzowlVideo}. Video identification attacks aim to infer videos that victims are watching without authorization. They pose an increasing privacy threat, as an individual's video viewing history or preference may reveal their political, financial, and personal interests~\cite{bae2022watching}, and others may judge them based on such data~\cite{frankowski2006you}, causing cyberbullying and discrimination. Moreover, scammers may use sensitive video viewing history to conduct blackmail, threatening to release it to victims' families, friends, coworkers, or social network contacts~\cite{CallawayThe,FazziniEmail}. In the United States, the Video Privacy Protection Act (VPPA)~\cite{CongressVideo} was enacted in 1988 after Robert Bork's video rental history was published during his Supreme Court nomination, making it illegal to disclose video viewing history without the watchers' consent~\cite{HavardThe}. 

A traditional and naive method to identify a video is video content matching (e.g.,~\cite{LeeRobust}). An adversary may steal frames or audio data from a video snippet (e.g., via shoulder surfing attacks or recording with a camera or microphone), using them to generate features (i.e., video fingerprints) to characterize the video clip. Such features are then matched with a database of video fingerprints built with known videos. Videos containing content similar to the snippet are thus identified. However, such attacks rely on direct views of the victim's screen by capturing a clear view of the video content or eavesdropping accompanying audio information.
 
Recent research focuses on developing video identification attacks by analyzing streaming video traffic. Each video generates a distinct traffic pattern due to its unique content. Such a mapping can be pre-built for inferring what video is currently streaming. Those attacks, however, often require the attacker to either compromise the router that the victim's device connects to~\cite{GuWalls,GuTraffic-Based,ZhangTraffic} or infect the victim's device with malware (e.g., via rogue websites~\cite{RoeiBeauty}) to capture streaming traffic. Both requirements impose practical hurdles, making such invasive attacks no longer viable, as modern networks usually adopt anti-malware tools and are also secured with a password unknown to the sniffer. Some attacks~(e.g., \cite{bae2022watching,LakshmananOn}) using passive network reconnaissance are non-invasive, while they require special equipment such as Universal Software Radio Peripheral (USRP) to measure traffic. Moreover, most video streaming services (e.g., Netflix~\cite{NetflixHow}) offer offline viewing, generating no traffic and failing all existing traffic analysis based techniques.

\begin{figure}[t]
\centering \includegraphics[width=3in]{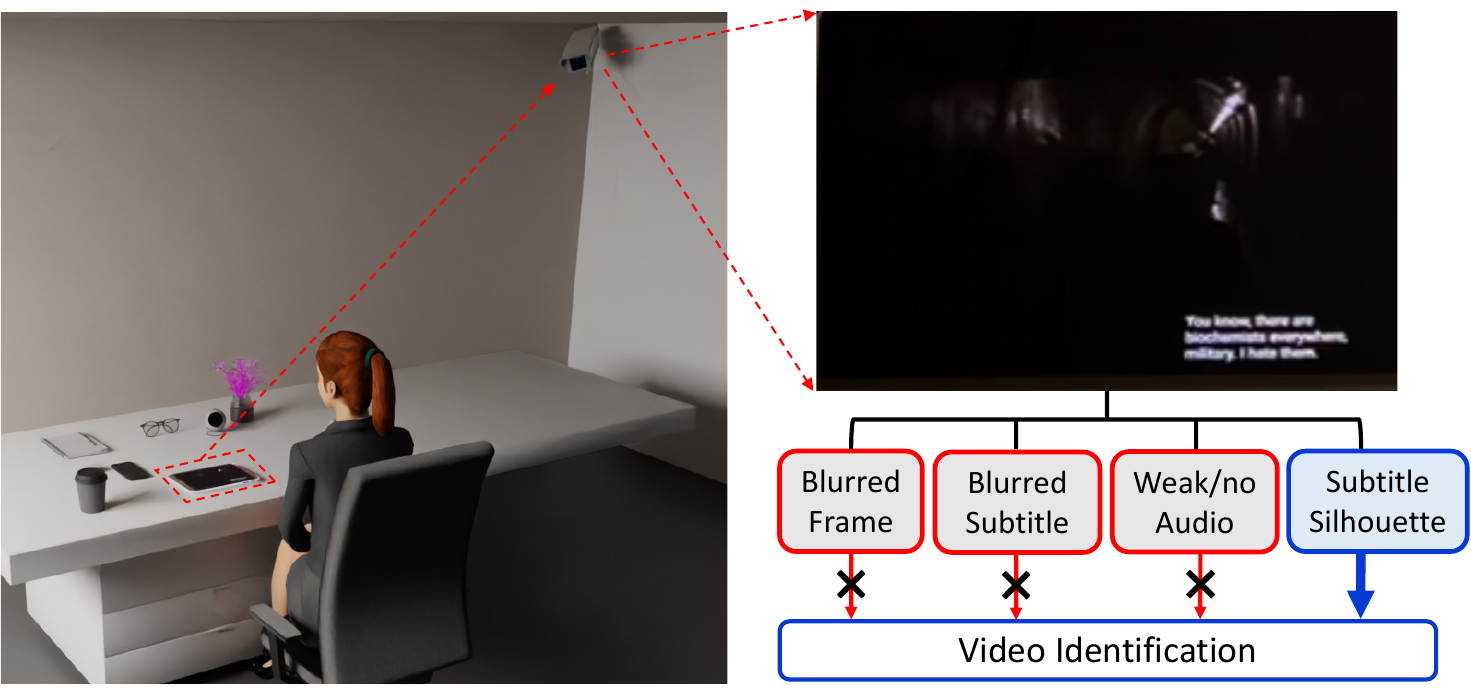} 
\vspace{-0.1in}
\caption{Subtitle silhouette information for inferring videos.} 
\label{fig:subtitle_motivation}
\vspace{-0.25in}
\end{figure}

Individuals often take precautions to protect video privacy, including preventing others from directly viewing the video content or hearing the audio, securing the video streaming network, or pre-downloading videos. In this paper, we investigate whether users are vulnerable to video identification attacks in general settings, after taking the aforementioned precautions. We discover it is possible to infer the video a victim is watching, whether online or offline, by pointing a single RGB camera at the victim's screen from a distance and obtaining the subtitle silhouettes of the video. We refer to the proposed attack as \textit{SilhouetteTell}. The silhouette (e.g., height or line count) of the video's subtitle on screen varies with the subtitle content. Such information can be captured by a camera at a distance, where the video frames, audio, and subtitles are completely illegible. The obtained coarse information related to subtitles can be exploited to infer videos, as shown in Figure~\ref{fig:subtitle_motivation}.

Subtitles are text representing the contents of the audio in a video, often displayed with each scene at the bottom of the screen. They are necessary for people with hearing impairments, facilitating following the dialog. Also, translated subtitles are necessary for media in a foreign language~\cite{HuSpeaker,cintas2014audiovisual}. According to a survey conducted in 2023, 63\% of young adults under 30 prefer watching videos with subtitles even in a language they know, and only 27\% choose subtitles off~\cite{BallardMost}. Another study of 5,616 participants shows that 69\% view video with sound off in public places and 25\% watch with sound off in private places, and instead, they turn on subtitles for understanding videos in these scenarios~\cite{McCueVerizon}.  
It has been well known that subtitles can be utilized to classify movie genres~\cite{RajputA}, improve movie recommender systems~\cite{EdenInvestigating}, and retrieve relevant videos from a large corpus~\cite{lei2020tvr}. However, in our general threat model, due to the long shooting distance and non-specific viewing angle, the attacker cannot record the content of the video that the victim is watching, nor can they obtain the subtitle content. Typically, when a user is watching videos that they wish to keep private, they tend to take actions, such as adjusting the screen away from the bystander. However, if others are too far away to see the screen, they often pay less attention and become less vigilant. 

Intuitively, Optical Character Recognition (OCR) algorithms may help recognize subtitles, by converting optically sensed document text (i.e., typed, handwritten, or printed text in images) into machine-encoded text~\cite{NagyTwenty}. We pick two most popular and state-of-the-art algorithms, Tesseract OCR Engine~\cite{SmithAn} and Convolutional Recurrent Neural Network (CRNN)~\cite{ShiAn}, which are deep learning-based and open-source. They both fail due to the poor recording quality. For the frame in Figure~\ref{fig:subtitle_motivation}, the subtitle is: ``\textit{You know, there are}'' (first line),  ``\textit{biochemists everywhere,}'' (second line),  and ``\textit{military. I hate them.}'' (third line). By running the latest Tesseract version 5.0~\cite{Tesseract} to this frame, the recovery result is ``\textit{vou eon few ow}" (first line), ``\textit{Scien}" (second line), and ``\textit{rata | hate Pam}" (third line). Also, with the current CRNN implementation~\cite{CRNNgithub}, the recognition output of the frame is nothing. Incorrect or empty extraction shows the inability of OCR methods to recover subtitles from blurry recordings.

We also wonder whether image captioning techniques, which generate natural language descriptions of images~\cite{HossainA}, can gather useful information for inferring videos. Vision-language pre-training (VLP) is currently the dominant training method using pre-trained large-scale models for visual recognition tasks~\cite{GhandiDeep}. One widely used pre-trained model is CLIP (Contrastive Language-Image Pre-Training)~\cite{RadfordLearning}. 
We utilize ClipCap~\cite{mokady2021clipcap}, a recent work based on CLIP and a pre-trained language model (GPT-2~\cite{radford2019language}), to generate captions for images. Similar to the challenges faced by OCR methods in recognizing blurry subtitles, it becomes evident that no correct caption can be produced for those blurry recordings. For example, in Figure~\ref{fig:subtitle_motivation}, the true caption is ``Two men are talking to each other", while the result of applying the ClipCap implementation~\cite{CLIP} is ``a scene from the movie", providing no help in video identification. 
 
{\color{black} Traditional ML-based techniques (e.g., OCR, CRNN, and ClipCap) have inherent limitations in recognizing subtitle or image content from blurry frames captured at a distance. In contrast, contentless subtitle silhouettes are much easier to distinguish under such conditions. Building on this observation, \textit{SilhouetteTell} extracts} spatial and temporal features for inferring videos.
First, we normally can observe a seemingly ``continuous white area'' in each video frame when a subtitle is present. We refer to this area as the ``subtitle area''. 
Different lengths of subtitles may result in subtitle areas with different shapes (silhouettes). This allows us to compare line counts among subtitles. Meanwhile, the duration of a subtitle may span multiple frames, that is, the subtitle remains unchanged while the video scene changes. We extract the line count of each subtitle as its spatial feature. Second, identifying subtitle areas for each frame helps derive the temporal difference between two successive frames. As aforementioned, the shape of the subtitle area varies with the subtitle content. However, for some subtitles with similar lengths, the shapes of their subtitle areas would be quite similar. For such cases, we observe that the overall brightness (i.e., the sum of image pixel values) of subtitle areas would differ as the subtitle contents change. Accordingly, we utilize both the shape and brightness of subtitle areas to distinguish whether the subtitle changes across neighboring frames. Such pattern variation of subtitle areas over time is referred to as \textit{temporal feature} of subtitles, which discloses how long (or how many frames) a subtitle appears on the screen.   

\textit{SilhouetteTell} initially needs to extract correct subtitle silhouettes in blurry videos, from where we barely identify any texts. 
As a blurry subtitle in videos recorded from a far distance often appears as a white chunk, we change the problem from recognizing texts to localizing white areas for obtaining subtitle areas, and design a subtitle silhouette extraction scheme using Mask Region-based Convolutional Neural Network (Mask R-CNN), a deep learning technique that can segment and identify the pixel-wise boundaries of each object~\cite{He_2017_ICCV}. Moreover, an attacker must compare the correlations among subtitles with those among observed subtitle silhouettes. This requires a self-contained spatiotemporal feature that can quantify such correlations and be compared against others. 
We create such a feature and corresponding approaches to compare observed subtitle silhouettes to possible candidate videos using this feature. Our technique incorporates a mechanism to retain high accuracy in the presence of subtitle silhouette recognition errors. 

In summary, we mainly make the following contributions.

\begin{itemize}[topsep=0pt, partopsep=0pt, parsep=0pt, itemsep=5pt, after=\vspace{-0.17in}]

\item We propose a new type of video identification attack using only video captured from a distance through commodity phone cameras. 
Our method does not require online video viewing, professional equipment, or a connection to the same network as the playback device.
\item We develop an algorithm to map obtained contentless subtitle silhouettes in recorded blurry videos into a video clip, and achieve video inference by modeling, extracting, and correlating their self-contained spatiotemporal features.
\item Extensive real-world experiments on top of 300 movies from YouTube, Amazon Prime Video, and Netflix, show that with a 2-minute blurry video clip, the average probabilities of recognizing videos in the top 1, 5, and 10 candidates can reach as high as 91.1\%, 98.7\%, and 99.8\%, respectively. 
\end{itemize}

\section{Preliminaries}
\label{sec:preliminaries}

\textbf{Subtitles:} A subtitle, in the form of one or more lines of written text, typically appears at the bottom center of the screen in sync with a video's audio~\cite{gottlieb2012subtitles,BrownDynamic}. It may appear in a rectangular overlay with a colored or opaque background. 

Subtitles can be open or closed. Open subtitles are embedded in the video and cannot be turned off, whereas closed subtitles are on a separate text track and can be turned on/off by the viewer. Subtitles can be downloaded in specific formats, e.g., SubRip text (.srt), Web Video Text Tracks Format (.vtt), and SubStation Alpha (.ssa)~\cite{zarate2021captioning}. The .srt format, as one of the most popular and almost a de facto standard in the web~\cite{zarate2021captioning,hanke2020extending}, contains the subtitle index, start time, end time, and text content. In a .srt file, subtitles are numbered sequentially, and the timecode format used is hours:minutes:seconds, milliseconds, with time units fixed to two zero-padded digits and fractions fixed to three zero-padded digits (00:00:00,000). 

\textbf{Mask R-CNN: } Object detection deals with the task of segmenting instances of semantic objects in digital images and videos. We input an image to an object detection model, which generally outputs coordinates of bounding boxes in the input image that contains specific objects. Faster R-CNN~\cite{RenFaster} makes a series of improvements on the initial object detection algorithm of Regions with CNN features (R-CNN)~\cite{GirshickRich}. Mask R-CNN is an extension of Faster R-CNN and augments object detection by adding object segmentation. 

\section{Threat Model and Assumptions} 
\label{sec:model}

We consider a general user (victim) who might be vulnerable while watching subtitled videos on computers/tablets/phones in open places such as libraries, airports, cafes, and cubicles. An adversary records the scenario where the victim is watching videos from a distance via a single RGB camera (e.g., a phone camera) and processes the recorded blurry video to identify videos the victim watches. An attacker is assumed to have access to the victim's space, either physically or remotely, to monitor their screen activity.

\textbf{No Traffic Access:} Our work differs significantly from prior work~(e.g., \cite{ReedIdentifying,GuWalls,LiDeep,DubinI,bae2022watching,LakshmananOn}) in that we do not rely on traffic information of video streaming sessions. 
\textit{SilhouetteTell} thus works even in scenarios where the victim watches videos offline, while all previous traffic analysis based video identification attacks fail. 

\textbf{No User-specific Training and Directly Observing Video Content:} We assume that the attack is opportunistic, requiring no labeled data or prior observation of the victim, and the victim is alert to conventional shoulder-surfing attacks in the sense that the attacker cannot get too close to them when watching a video.

\textbf{Recording Required: } While directly observing an individual may sometimes reveal sensitive information (e.g., their lifestyle), such inferences typically require prolonged observation and may be ineffective if the victim does not outwardly exhibit personal traits. The attacker does not need her recordings to contain the full screen of the target device on which the victim watches the video, while we assume that the attacker can observe at least a partial blurry view of the screen containing the subtitles. Such scenarios are quite common in practice as it is challenging for the victim to block observing the screen from all angles.

\textbf{Subtitle Library:} We also assume that the attacker has access to the subtitles of all videos in the suspect set. Such knowledge can be easily acquired as subtitle files are often available and downloaded from subtitle websites (e.g.,~\cite{Opensubtitle24}) or video streaming services. Appendix~\ref{appendix:crawling} discusses specific methods for crawling subtitle files.

\section{Video Inference Attack} 
\label{sec:video_inference}

\begin{figure}[t]
\centering \includegraphics[width=3.2in]{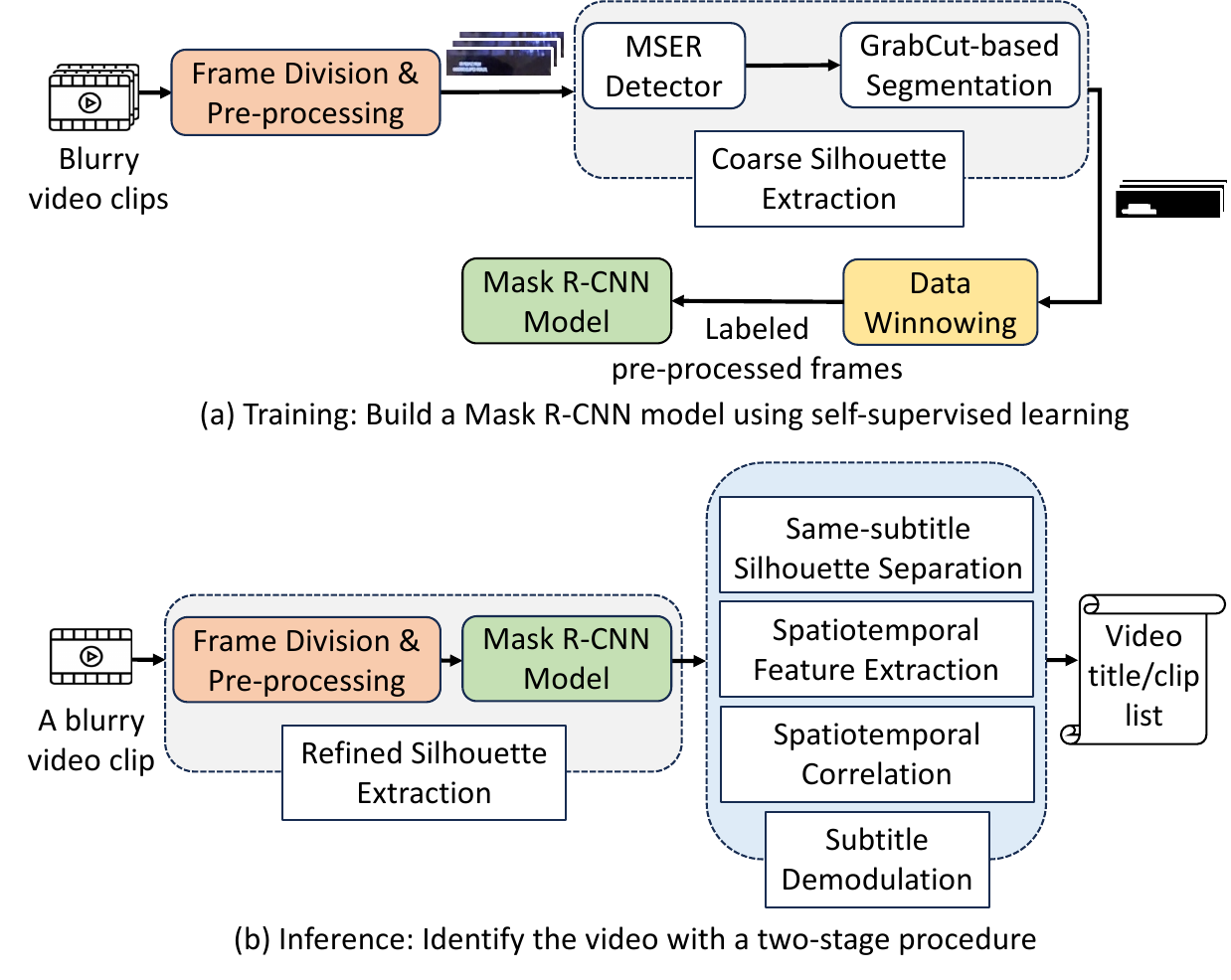}
\vspace{-0.15in}
\caption{Overview of \textit{SilhouetteTell}.}
\label{fig:overview}
\vspace{-0.15in}
\end{figure}
\subsection{Attack Overview} 
\label{subsec:overview}

\textit{SilhouetteTell} performs a two-phase process to infer videos from observed blurry videos: \textit{training} and \textit{inference}. 

Figure~\ref{fig:overview} plots an overview of \textit{SilhouetteTell}. Figure~\ref{fig:overview}a depicts the offline training phase, where a Mask R-CNN model is trained for accurately detecting subtitle silhouettes. The attacker first divides the recorded video into frames, and pre-processes them to focus on the subtitle areas on the victim's screen. The pre-processed frames are input into a module of \textit{Coarse Silhouette Extraction}, built on Maximally Stable Extremal Regions (MSER)~\cite{MatasRobust} and GrabCut~\cite{RotherGrab}, to obtain raw training data,  which are then sanitized via a module of \textit{Data Winnowing}. Figure~\ref{fig:overview}b shows the inference phase, consisting of two modules, \textit{Refined Silhouette Extraction} and \textit{Subtitle Demodulation}. For the initial module, the attacker obtains pre-processed frames using the same procedure in the training phase, and inputs them to the trained Mask R-CNN model. In the second module, we separate the captured silhouettes originating from different subtitles, and obtain a silhouette sequence. Each element of the sequence belongs to a corresponding distinguishable subtitle. Next, we extract the spatiotemporal feature of the silhouette sequence, and correlate it with a subtitle sequence presented in subtitle files, shrinking the candidates of the target video title and clip.

\subsection{Training Phase}
\label{subsec:training}

\subsubsection{Frame Division \& Pre-processing}
We divide the recorded video clip into individual raw video frames using OpenCV, an open-source library for computer vision~\cite{OpenCV24}. A frame may record the background beyond the victim's screen, where certain video scenes may confuse the subtitle area recognition. Accordingly, we perform a dual-cropping operation on each frame uniformly to eliminate interference and only include the area where subtitles may appear. 
 
\textit{First-cropping:} 
If the victim's screen is not moved in the recording, we can mark the screen's four corners to designate the portion of the video frame and clip the corresponding area. If the screen's location varies in the recording (e.g., when the recording camera is not stationary), the manual cropping may be laborsome. Alternatively, we use an existing object detection method (i.e., a pre-trained Mask R-CNN model~\cite{Mask_RCNN_2024}) to crop out the screen area efficiently. 

\textit{Perspective Transformation:} We may record the screen from different angles. Thus, the subtitle area in the raw recording may be skewed or compressed. To reduce such effects, we apply perspective transformation to convert an original screen recording image into the plane of the screen (i.e., the bird’s eye view) before applying the second cropping. Specifically, we first mark the 4 points on the video to indicate the target screen's planar surface. We later compute a homography matrix $\mathbf H$ between this planar surface and the video frame's perspective, and finally transform the source image with the matrix $\mathbf H$, using two OpenCV functions, $perspectiveTransform$ and $warpPerspective$~\cite{Geometric}. Let $(x,y)$ denote the 2D coordinate of a point in the source image. By multiplying $(x,y)$ with $\mathbf H$, we obtain its corresponding point in the plane of the screen.
 
\textit{Second-cropping:} 
Occasionally, the screen may have regions that appear similar to a subtitle area. Minimizing such similarity would be beneficial for recognizing subtitle areas. Empirically, we observe that the subtitles of videos for popular video streaming services (e.g., YouTube, Netflix, and Amazon Prime Video) are usually displayed in the bottom quarter of the screen. We then crop this area as the input to the next step for subtitle silhouette extraction. If subtitles appear in other parts of the screen, the second cropping will target that area, adapting to corresponding subtitle displacements.
 
\subsubsection{Coarse Silhouette Extraction} 
This step aims to extract subtitle silhouettes leveraging the MSER feature extractor and GrabCut-based segmentation. Such a process still suffers from segmentation errors, and we refer to it as coarse silhouette extraction. 

\textit{MSER Feature Extractor:} In the pre-processed frames, normally, the subtitle area and non-subtitle area (i.e., background) differ in color, while within each area, the color is nearly constant. MSER is a blob detector that finds the stable connected regions in a gray-scale image over a wide range of thresholds. To transfer an RGB image into a gray-scale one, we utilize an OpenCV function $cvtColor$~\cite{color24}, which represents each pixel in the original image with a single grayscale intensity according to its RGB values. The intensity is denoted with an 8-bit integer giving 256 possible different shades of gray from black to white. After applying MSER to each pre-processed frame (with the grayscale format) with another OpenCV function - $MSER\_create$~\cite{MSER24}, a sequence of rectangular bounding boxes will be obtained. Next, for each bounding box, we calculate the corresponding edge-to-edge distances of the pre-processed frame and this box, resulting in four distances. If any of the distances is smaller than an empirically pre-defined threshold, we determine that this box is near the edge of the pre-processed frame, and then filter it out, as it usually may not contain the subtitle. 

%Furthermore, t
To select the most relevant bounding box corresponding to the subtitle area, we propose a bounding box selection algorithm based on the Non-Maximum Suppression (NMS) method~\cite{neubeck2006efficient,hosang2017learning}. Let $A$ and $B$ denote two bounding boxes. Their Intersection over Union (IoU) is the ratio between the area of overlap to the area of union between $A$ and $B$, i.e., $\frac{A \cap B}{A \cup B}$. We use the standard IoU threshold of 0.5~\cite{hosang2017learning} and perform the following steps. First, a confidence score $\mu$ for each bounding box is defined as the ratio of its height to its width. We sort the boxes in descending order of $\mu$, and denote the one with the highest $\mu$ as $H$. Empirically, we find $H$ usually recognizes the object more accurately than the rest. Next, we compute $IoU$ of $H$ with every remaining box. Let $C$ denote one of the remaining boxes. If $IoU$ of $H$ and $C$ is greater than 0.5, we remove $C$. After traversing all boxes, if it ends up with two or more unique ones, we observe this case often appears when there are some small bounding boxes encapsulating inference of white points, rather than the subtitle. Accordingly, we keep the one with the largest size. 

\textit{GrabCut-based Segmentation:} The next step is to obtain subtitle silhouettes from the bounding boxes. GrabCut is often used to segment the foreground of an image from the background, while it is required to first manually frame and select the target area~\cite{RotherGrab}. Instead, we utilize the module of the MSER feature extractor to automatically generate the input of GrabCut, replacing the low-efficiency process of manually choosing the segmentation range. 
Generally, for each bounding box, the segmentation results using an OpenCV function - $grabCut$~\cite{grabcut24}, return the subtitle silhouette as the foreground and the rest as the background.

\subsubsection{Data Winnowing} 
The results of coarse silhouette extraction may contain errors. 
We thus perform a consistency check to see whether the segmented subtitle silhouettes are correct. Particularly, we winnow out the pre-processed frames, whose subtitle silhouettes are incorrectly recognized. 
Meanwhile, the rest pre-processed frames, together with their corresponding detected subtitle silhouettes, become training data for building a Mask R-CNN model. 

\subsubsection{Mask R-CNN Model Building} 

We utilize a commercial smartphone (Samsung Galaxy Z Fold4) to record a laptop playing movies on three streaming services (YouTube, Amazon Prime Video, and Netflix) at distances larger than 4 meters. In these recordings, both the movie and the subtitle contents appear blurry.  

\textit{Dataset Splitting: } 
For each streaming service, we select 10 popular movies, and for every movie, we randomly record a 2-minute clip with 60 FPS for processing. As a result, we have a total of $3 \times 10 \times 120 \times 60 =216,000$ raw video frames as input. The coarse silhouette extraction recognizes 109,326 frames that contain subtitle areas. After applying data winnowing, we obtain 78,591 pieces of correctly labeled data (referred to as \textit{silhouette-contained data}). Also, among falsely labeled data, we obtain 3,244 pieces, in each of which the corresponding pre-processed frame actually has no subtitle silhouette. For such data, we replace the incorrect original label with a segmentation result of no subtitle silhouette, and refer to these revised data as \textit{silhouette-free data}. {\color{black} The only component that requires training is the Mask R-CNN model.} With the two obtained datasets, silhouette-contained and silhouette-free, we split each of them using the common 80:20 train-test ratio for fine-tuning a Mask R-CNN model. We use ResNet50~\cite{he2016deep,He_2017_ICCV} as the backbone and stochastic gradient descent (SGD)~\cite{bottou2010large} for optimization.

\begin{figure}[t]
\centering \includegraphics[width=2.6in]{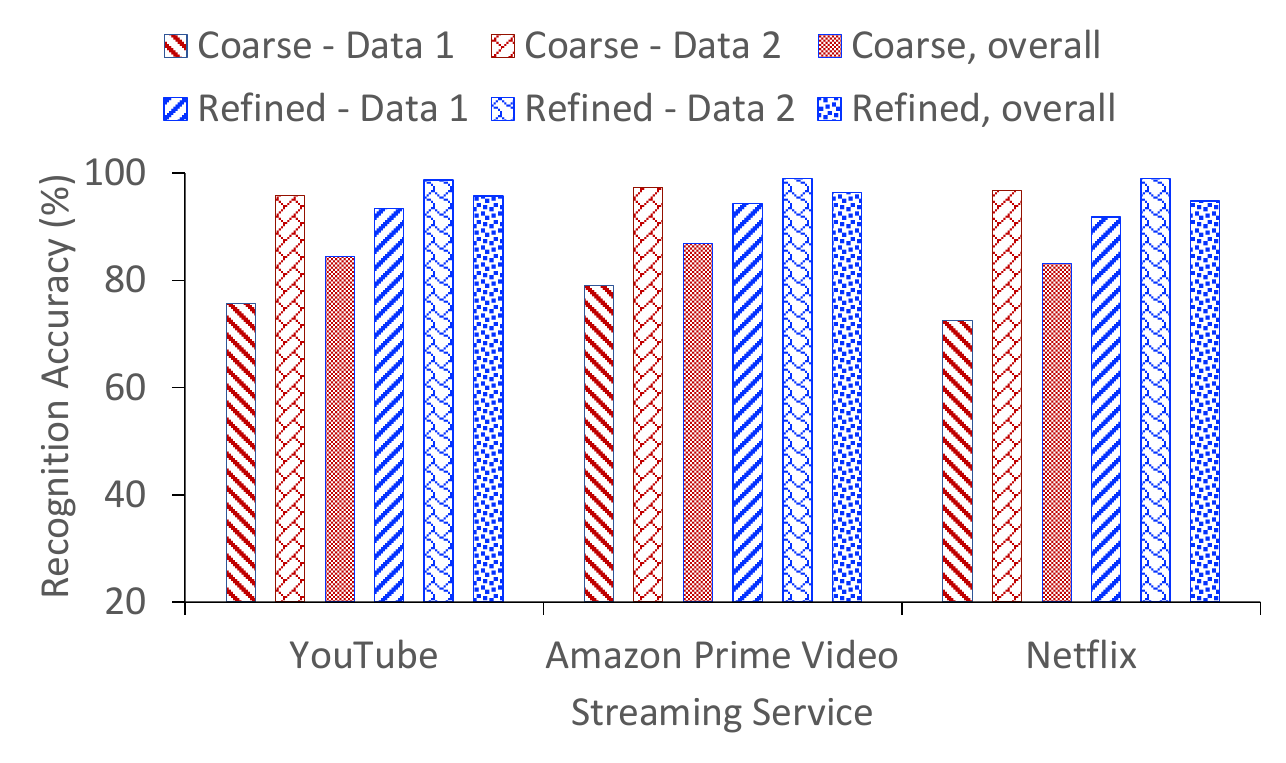}
\vspace{-0.15in} 
\caption{Performance of coarse and refined silhouette extraction (denoted as ``Coarse" and ``Refined") for silhouette-contained (Data 1) and silhouette-free frames (Data 2).}
\label{fig:coarse_refined_cmp} 
\vspace{-0.05in}
\end{figure}
\subsection{Inference Phase}
\label{subsec:inference}

\subsubsection{Refined Silhouette Extraction}
\label{subsubsec:refined}

Compared with coarse silhouette extraction, refined silhouette extraction can achieve higher accuracy. We verify this with a new dataset. For this dataset, we randomly select 15 new movies (not shown in the training dataset) from the three streaming services, with 5 for each. We then record a randomly picked 2-minute video for every movie with 60 FPS. In total, we have $15\times 2 \times 60 \times 60 \!=\! 108,000$ frames. {\color{black} This dataset is only used to validate the superior performance of the refined silhouette extraction method over the coarse one.
}

Figure~\ref{fig:coarse_refined_cmp} compares the recognition rates of coarse and refined silhouette extraction methods applied to silhouette-contained and silhouette-free frames. We can observe that the refined silhouette extraction method consistently achieves higher recognition accuracy compared to the coarse one. Also, it always maintains high recognition accuracy regardless of the streaming service or whether the input frame has a subtitle. Specifically, the overall recognition accuracy for refined silhouette extraction is consistently above 95\%, whereas for coarse silhouette extraction, this value is consistently below 86\%. These results convincingly demonstrate the necessity of building a Mask R-CNN model for recognizing subtitle silhouettes. Also, Appendix~\ref{appendix:com} shows example cases where refined silhouette extraction works while coarse silhouette extraction fails.

\subsubsection{Subtitle Demodulation}
\label{subsubsec:sub_demo}

Subtitle demodulation converts a sequence of consecutive subtitle silhouettes into corresponding subtitles of a video, enabling video identification. 
We start by distinguishing silhouettes originating from the same subtitle and designing a spatiotemporal feature to be applied to a subtitle silhouette sequence. This feature must be suitable for narrowing down the search space of possible candidates. Subsequently, we demonstrate how to apply this feature to multiple subtitle silhouette sequences.

\textit{Same-subtitle Silhouette Separation:} 
\begin{figure}[t]
\centering \includegraphics[width=2.6in]{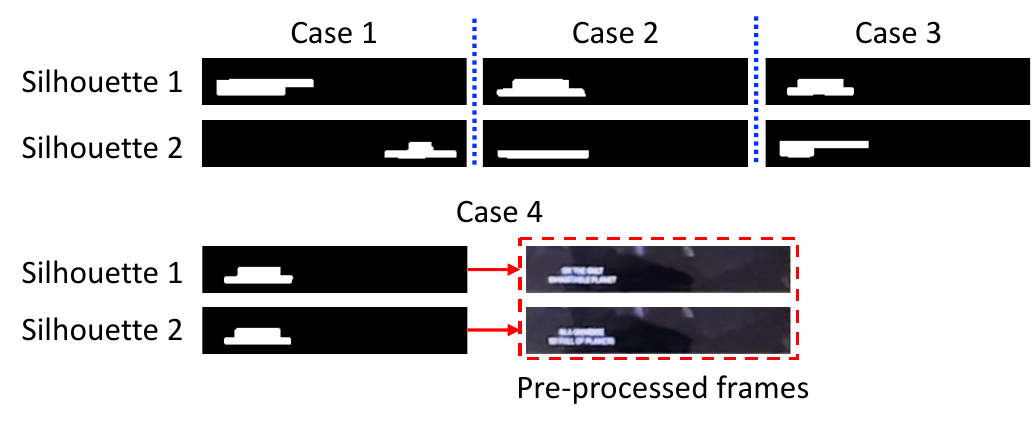}
\vspace{-0.15in}
\caption{Silhouette comparison: entirely separate (case 1); partially overlap with varying or the same line counts (case 2/3); fully overlap and check pre-processed frames (case 4).} 
\label{fig:cross-frame}  
\vspace{-0.15in} 
\end{figure} 
The same subtitle usually generates highly similar silhouettes at the same positions of a frame. If two silhouettes are different, they normally come from different subtitles. Figure~\ref{fig:cross-frame} presents typical scenarios involving two consecutive silhouettes. Intuitively, it is easy to determine that in the first three cases, the two silhouettes, which either have no overlap or partially overlap, come from different subtitles. Specifically, to reduce the impact of the no-silhouette area, we place two marked silhouettes $S_1$ and $S_2$ into the same frame, and then compute their IoU, which is denoted as $IoU_{s}= \frac{S_1 \cap S_2}{S_1 \cup S_2}$, ranging from 0 to 1. 

\begin{figure}
\hspace{0.03in}
\begin{minipage}[t]{0.475\linewidth}
\centering
\includegraphics[width=1.65in]{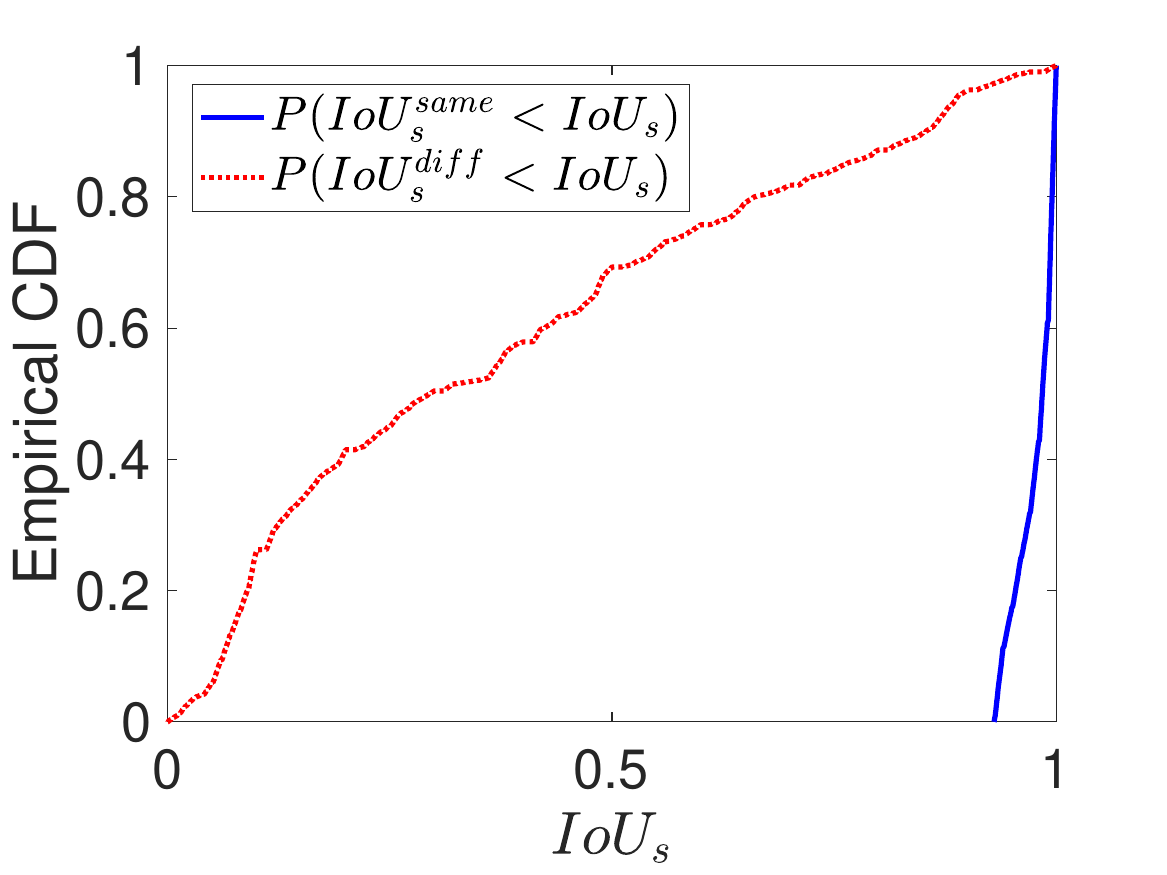} 
\vspace{-0.3in}
\caption{CDFs of IoU of two silhouettes.} 
\label{fig:cdf_IoU}
\end{minipage}
\hspace{0.03in}
\begin{minipage}[t]{0.475\linewidth}
\centering
\includegraphics[width=1.65in]{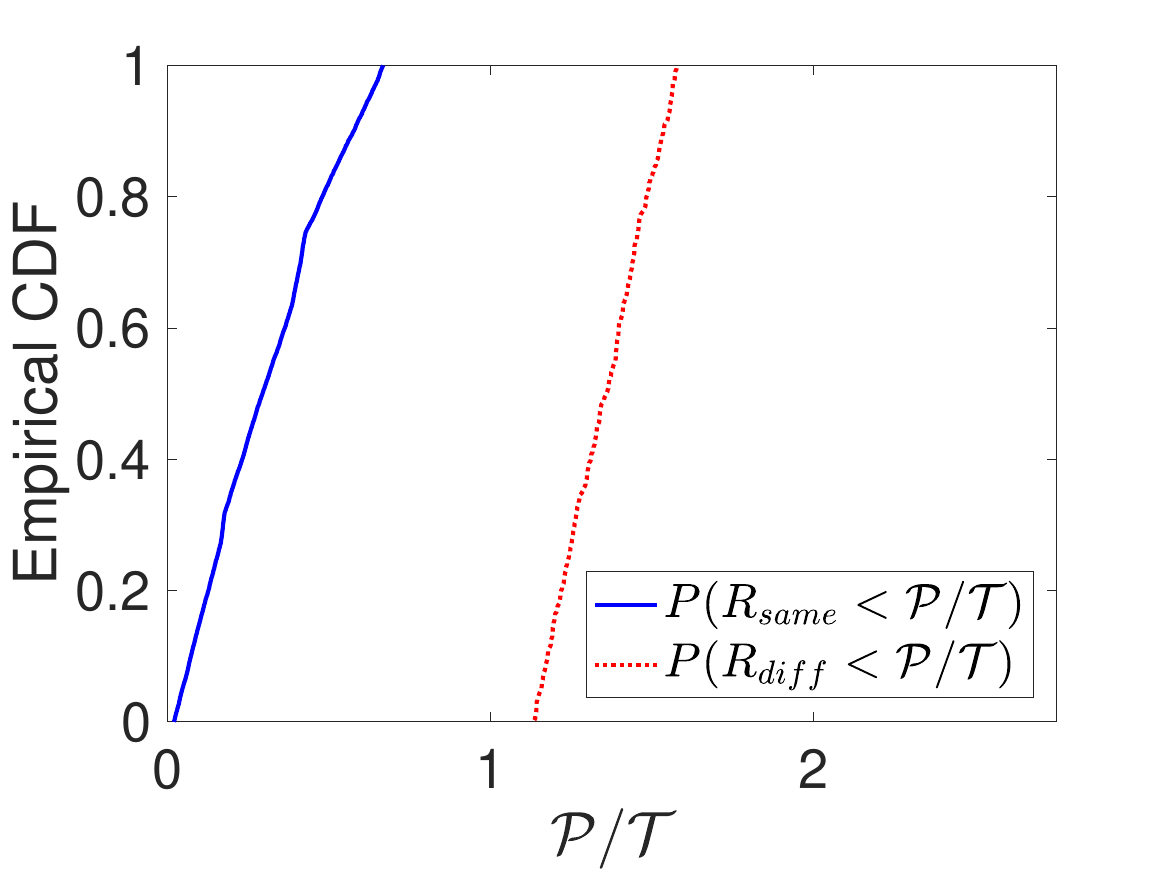} 
\vspace{-0.3in}
\caption{CDFs of similarity ratio of two silhouettes.} 
\label{fig:pt_ratio}
\end{minipage}
\vspace{-0.3in}
\end{figure}

Let $IoU^{diff}_{s}$ and $IoU^{same}_{s}$ denote the IoU of two neighboring silhouettes coming from different and the same subtitles. To explore their practical values, we randomly select 10,000 pairs of neighboring silhouettes coming from the same and different subtitles, respectively. Figure~\ref{fig:cdf_IoU} plots the empirical cumulative distribution functions (CDFs) of $IoU^{diff}_{s}$ and $IoU^{same}_{s}$. We see that $IoU^{same}_s$ is always near 1 (above 0.93), while $IoU^{diff}_{s}$ is less than 0.93 with a probability of 97\%. Thus, with an IoU threshold of 0.93 for distinguishing silhouettes, the true positive (i.e., silhouettes originating from the same subtitle are correctly identified) and true negative (i.e., silhouettes from the different subtitle are correctly identified) would be 100\%, while the false positive (silhouettes from different subtitles are incorrectly identified as from the same subtitle) is 3\%. This appears as the IoU-based method cannot distinguish between two silhouettes from different subtitles when they have similar sizes and appear at similar locations. For example, in case 4 of Figure~\ref{fig:cross-frame}, both silhouettes are similar with IoU equaling 0.95, while the corresponding subtitles differ. We further compare perspective pre-processed frames associated with them. This is based on the observation that different subtitles have distinct words and sentence structures, leading to variation in pixel value distribution in subtitle areas (i.e., difference in pre-processed frames). 

When the IoU of two neighboring silhouettes is above or exceeds the threshold, we first put two marked silhouettes into the same frame, and identify the area of intersection. We then extract the portion marked by this intersection area in each of the two corresponding pre-processed frames. We refer to the extracted two regions as $A_1$ and $A_2$, respectively, and calculate the absolute difference between every pixel in $A_1$ and the corresponding pixel in $A_2$. Finally, we sum all such pixel differences and denote the sum as $\mathcal P$. We set up another {threshold} $\mathcal T=\mathcal T_0 \cdot N$, where $N$ is the number of pixels in $A_1$ or $A_2$ and $\mathcal T_0$ indicates the mean maximum allowable pixel variance along two neighboring frames for the same subtitle. 
We introduce a new metric, called \textit{similarity ratio}, and denoted as $R=\frac{\mathcal P}{\mathcal T}$ to compare $\mathcal P$ and $\mathcal T$. If $R>1$, we have $\mathcal P> \mathcal {T}$, and regard that the silhouettes originate from different subtitles; otherwise, when $0\leq R \leq1$, they are determined as from the same subtitle. 

Empirically, we choose $\mathcal T_0 = 1.5$, achieving high accuracy. We also pick another 10,000 pairs coming from different subtitles while each such pair has an IoU larger than 0.93. Figure~\ref{fig:pt_ratio} presents the CDFs of $R_{same}$ and $R_{diff}$ that denote the similarity ratio for a pair of neighboring silhouettes coming from the same and different subtitles, respectively. We can see that $R_{same}$ is always below 1 while $R_{diff}$ is consistently above 1, indicating a 100\% success rate in distinguishing silhouettes originating from the same or different subtitles.
Particularly, for case 4 in Figure~\ref{fig:cross-frame}, we have $\mathcal T= 17,145$, and $\mathcal P = 20,298  > \mathcal T$, where the two silhouettes are correctly determined as coming from different subtitles.

\textit{Spatiotemporal Feature Extraction: }  
Ideally, the feature extracted from a subtitle silhouette sequence would enable us to uniquely determine the video clip (and thus the video title). Suppose the duration of the recorded clip is $T$ and we can accordingly extract $C$ clips with a period of $T$ within the suspect library of videos. A perfect feature would classify the $C$ video clips into $C$ groups, each having one member only, such that an input subtitle silhouette sequence can uniquely match a video clip based on this feature. Our strategy is thus to figure out a feature that can divide all video clips into as many groups as possible, to achieve high distinguishability.
 
To quantify the distinguishability of a feature in dividing all candidates, we define \textit{uniqueness score} as a new metric, as the ratio $C_s/C_0$, where $C_0$ is the number of considered video clips and $C_s$ represents the number of sets obtained by dividing $C_0$ video clips with the selected feature. The uniqueness score should be maximized for the best partitioning of the video clips. To calculate the number of sets divided by each, we randomly select 10,000 different clips with a duration of $T$ from 100 movies streaming on YouTube, Amazon Prime Videos (Amazon for short), and Netflix respectively. We vary $T$ from 1 to 3 minutes, in increments of half a minute.

\textbf{Temporal Feature Representation: } From a temporal perspective, the intuitive feature of a video clip is the number of subtitles displayed, i.e., the length of the subtitle silhouette sequence. The sequence length can be obtained by counting the number of varying subtitle silhouettes within the period of the video clip since two successive frames showing the same subtitle would exhibit the same subtitle silhouettes. We find that all selected 2-minute 10,000 video clips for Netflix are 1-94 subtitles long. If we choose the sequence length as the only feature, we can divide all suspect Netflix video clips into 94 sets. On average, each set has $10,000/94 \approx 106.4$ video clips. The uniqueness score is then $94/10,000 =0.0094$.

\textbf{Spatial Feature Representation: } From a spatial perspective, the shapes of subtitle silhouettes disclose how they differ in terms of their width or height. 
Normally, the video subtitles have no more than three lines, avoiding obstructing the video scene. Also, the height of a subtitle silhouette is proportional to the line count. Thus, we use it to characterize each subtitle silhouette. Silhouettes with similar heights are clustered in the same group. Next, we sort and index all groups in ascending order of their heights. We denote the height similarity information of a silhouette sequence as $[s_{1}, \cdots, s_r]$, where $r$ is the number of distinct silhouette height groups that appear, and $s_{i}$ ($i\in \{1, \cdots, r\}$) denotes how many times a silhouette in the group with the $i^{th}$ smallest index appears. 

Considering that a subtitle has up to three lines, we have up to three categories according to the silhouette height. For instance, for a sequence of 5 subtitles whose lines are 2, 2, 1, 2, and 1, respectively, its height similarity information is [2, 3], as there are two different line counts, with the smaller one appearing twice and the other appearing three times. Using the height similarity information, we can divide all selected 10,000 2-minute Netflix video clips into a total of 789 sets. The uniqueness score equals $ 789/10,000 = 0.0789$.
 
 \textbf{Spatiotemporal Feature Characterization: }
The height similarity information has better distinguishability than the length feature, as its larger uniqueness score yields a smaller set cardinality, and hence a reduced search space to map an input silhouette sequence to a video clip. The feature of height similarity information gives the statistics of subtitles with varying lines/heights, while it does not consider their exact positions in the sequence. We expect a more advanced feature can further increase the uniqueness score by indicating both the length and height similarity information of the silhouette sequence, as well as the height variation from silhouette to silhouette. Let $\mathbf s=[s_1, s_2, \cdots, s_n]$ denote a sequence of $n$ different subtitles or subtitle silhouettes within a video clip of a period of $T$. We define its \textit{spatiotemporal vector} as 
 \begin{equation}
 V: \mathbf s=[s_1, s_2, \cdots, s_n] \longmapsto  \mathbf l=[l_1, l_2, \cdots, l_n],
 \end{equation}
where $l_i$ ($i \in \{1, 2, \!\cdots\!, n\}$) denotes the mapping of $s_i$ in the spatiotemporal vector. To construct $V$, the following steps are performed. 
\begin{itemize} 
\item We first cluster $n$ subtitles or silhouettes based on the line count of subtitles or the silhouette's height, and thus obtain $r$ sets.
Subtitles with the same line count, or silhouettes with comparable height, aggregate into a separate set.  
\item Each set is associated with a line count or height as a label. Sort $r$ sets based on this label and index them from 1 to $r$. 
\item Finally, $s_i$ ($i \in \{1, 2, \cdots, n\}$) will be mapped into the index of the set that $s_i$ belongs to, i.e., $l_i$. 
\end{itemize}

We build the spatiotemporal vector for each video clip and ultimately partition the 10,000 2-minute Netflix video clips into 9,271 sets. The corresponding uniqueness score is $9,271/10,000 = 0.9271$, much larger than those of the previously discussed two features.

\begin{figure}[t]
\centering \includegraphics[width=3.1in]{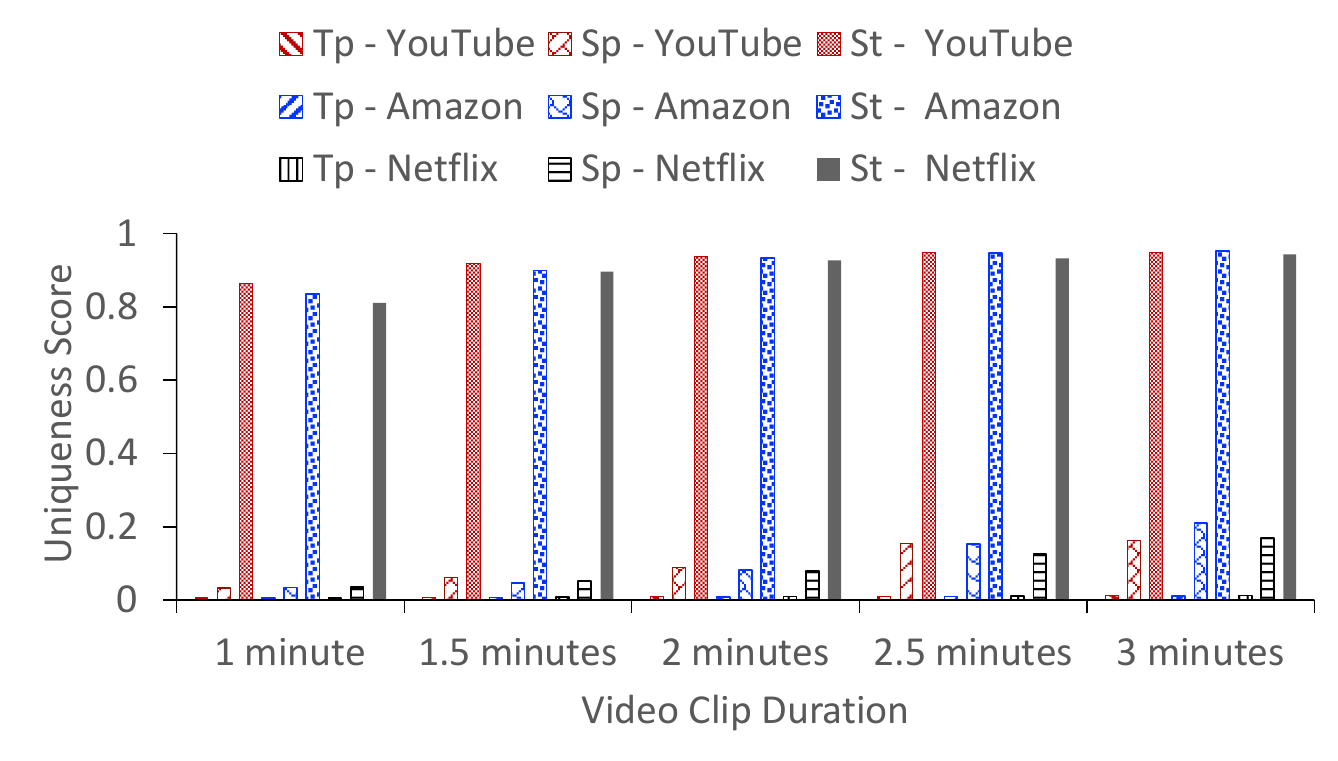} 
\vspace{-0.15in} 
\caption{Uniqueness scores of video clips.} 
\label{fig:score_cmp}
\vspace{-0.2in}
\end{figure}
Empirically, we find that the uniqueness scores for video clips of varying durations are not evenly distributed. Figure~\ref{fig:score_cmp} compares the uniqueness scores when we search candidates using sequence length (i.e., temporal feature or ``Tp''), height similarity (i.e., spatial feature or ``Sp''), and spatiotemporal vector (i.e., spatiotemporal feature or ``St''). For all cases, the spatiotemporal vector greatly outperforms the other two. There is no significant difference in uniqueness scores among different streaming services. However, it is evident that as the video clip becomes longer, it becomes more uniquely structured, resulting in a higher uniqueness score for each feature. For example, with the spatiotemporal vector as the feature to divide chosen Netflix videos, the uniqueness score for a one-minute video clip is 0.81, while that for a three-minute video clip reaches 0.95. Across all streaming services, video clips of 2 minutes or longer, maintain uniqueness scores above 0.93. 

\begin{figure}[t]
\centering \includegraphics[width=2.8in]{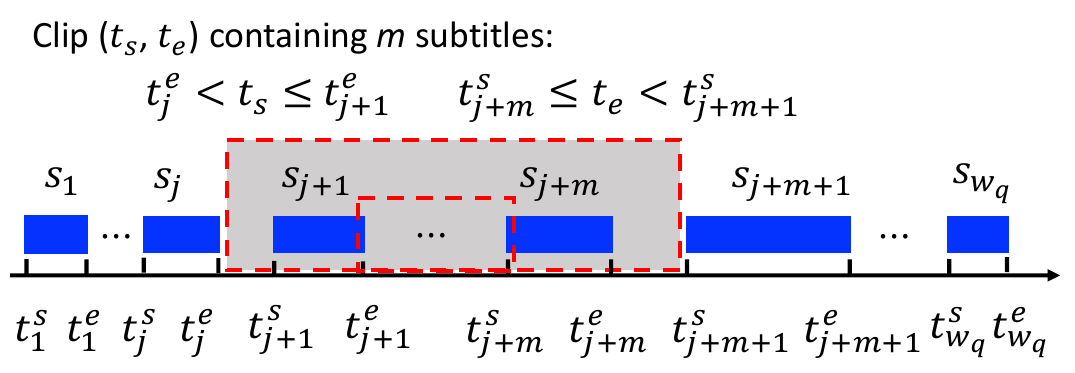}
\vspace{-0.1in} 
\caption{A sliding window of $m$ subtitle.} 
\label{fig:timeline}
\vspace{-0.26in}
\end{figure}
\textit{Spatiotemporal Correlation: } 
Suppose the suspect library has $K$ videos. With the subtitle file of the $q^{th}$ ($q \in \{1,2, \cdots K\}$) video, we can obtain its subtitle sequence $[s_1, s_2, \cdots, s_{w_q}]$, which $w_q$ denotes the video's total amount of subtitles, as well as the starting and ending time points of each subtitle, i.e., $t_i^{s}$ and $t_i^{e}$ ($i \in \{1, 2, \cdots, w_q\}$). For a recorded blurry video clip with a duration of $T$, its subtitle silhouette sequence is denoted with $\mathbf S=[S_1, S_2, \cdots, S_m]$, where $m$ denotes the total number of silhouettes. We can then obtain its spatiotemporal vector as $\mathbf V_{o}=V(\mathbf S)$.
Let $\mathcal C$ represent the demodulated set containing the candidates for this recording.

We utilize a sliding window of $m$ subtitles, with a step size of one subtitle, applied to the entire subtitle sequence of each video. Thus, the $q^{th}$ video will be divided into $w_q-m+1$ segments, and each can be denoted with $\mathbf G=[s_{j+1}, s_{j+2}, \cdots, s_{j+m}]$ ($j \in \{0, 1, \cdots, w_q-m\}$). Figure~\ref{fig:timeline} illustrates how we extract a segment with $m$ subtitles, i.e., $s_{j+1}, \cdots, s_{j+m}$. The starting and ending time points are denoted with $t_s$ and $t_e$. Accordingly, we obtain the following two constraints for them, $t_j^{e}<t_s \leq t_{j+1}^e$, and $t_{j+m}^{s} \leq t_e \leq t_{j+m+1}^s$. Particularly, we let $t_0^e$ and $t_{w_q+1}^{s}$ be 0 and $T$, indicating the start and end of the video, respectively. Beginning with searching from the first suspect video, we perform the following steps to decode the recorded blurry video. Initially, we set $q=1$, $j=0$, and $\mathcal C=\emptyset$. 

\begin{itemize}
\item[(1)] When the time constraints for the recording are not satisfied, i.e., $t_{j+m+1}^s-t_{j}^{e}<T$ or $t_{j+m}^s-t_{j+1}^{e}>T$, this $m$-subtitle window has no way to exhibit an $m$-silhouette sequence of duration $T$. We will thus skip this no-candidate segment. 
\item[(2)] When the two time constraints are satisfied, we calculate the spatiotemporal vector of $\mathbf G$ as $\mathbf V_c=V(\mathbf G)$. If $\mathbf V_c$ matches with $\mathbf V_o$, this segment $\mathbf G$ would be a candidate and we add it into the set $\mathcal C$; otherwise, we skip this segment.  
\item[(3)] If $j$ equals $w_q$, jump to step (4); otherwise increase $j$ by 1, and jump to step (1). 
\item[(4)] If $q<K$, we increment $q$ by 1, reset $j=1$, and jump to step (1); otherwise, return.  
\end{itemize} 
With the video clip candidates in $\mathcal C$, we can then retrieve their video titles to achieve video identification.  

\textbf{Joint Demodulation: } While watching a video, the victim may move their body to cover partial or all screen, leading to the attacker not recording the subtitle area on the victim's screen. Also, a pedestrian or obstacle may block the line-of-sight between the attacker and the victim's screen. Under these circumstances, the attacker may miss recording certain subtitle silhouettes, resulting in obtaining several discontinuous clips. In a general case, we assume that the attacker obtains $M$ clips, denoted as $\mathcal A= [\mathbf S_1, \mathbf S_2, \cdots, \mathbf S_M]$. The duration of these $M$ clips are represented by $T_1$, $T_2$, $\cdots$, and $T_M$. Also, the duration between two neighboring clips (referred to as the inter-sequence interval) is denoted by $T_{i, i+1}$ ($i \in \{1, 2, \cdots, M-1\}$). Our goal is to find a long video clip with duration $\sum_{i=1}^{M-1}(T_i+ T_{i,i+1})+ T_M$ that corresponds to the $M$ subtitle silhouette sequences. While each silhouette sequence could have several candidate short video clips with a matching structure, the combination of respective candidates for neighboring clips should satisfy the following two requirements: (1) both candidates come from the same video; (2) the interval between the two candidates within the video should match the inter-sequence interval observed by the attacker. Accordingly, the attacker first finds initial candidate short video clips for each subtitle silhouette sequence $ \mathbf S_j$ ($j \in \{1, 2, \cdots, M\}$), denoted as $\mathbf G_{j}=[\mathbf G_{j}^1, \mathbf G_{j}^2, \cdots, \mathbf G_{j}^{N_j}]$, where $N_j$ denotes the total amount of the candidates. The starting and ending time points for $\mathbf G_{j}^n$ ($n \! \in\! \{1, 2, \!\cdots\!, N_j\}$) are denoted with $T^s_{j}$ and $T^e_{j}$. The attacker then iterates over all $M$ clips with $j=2$ initially, and performs the following steps, returning when $j>M$. 
\begin{itemize}
\item[(i)] Concatenate the candidates for $\mathbf S_{j-1}$ and $\mathbf S_{j}$, obtaining $N_{j-1} \cdot N_j$ candidates. 
\item[(ii)] For each concatenated candidate, if $ \mathbf G_{j}^{n}$  and $\mathbf G_{j-1}^{m}$ ($m \!\in\! \{1, 2, \!\cdots\!,\\ N_{j-1}\}$) come from two different videos (i.e., with varying video titles), this candidate will be discarded;  
otherwise,  we further compare $T_{j-1,j}$ and $T_{j}^s-T_{j-1}^e$.
\item[(iii)] If $|T_{j-1,j}- (T_{j}^s\!-\! T_{j-1}^e)|\!>\! \delta$, where $\delta$ denotes the sum of the durations of the last and first subtitles of $\mathbf G_{j-1}$ and $\mathbf G_{j}$, respectively, the candidate will also be discarded. 
\item[(iv)] Update $\mathbf S_{j-1}\!=\! \mathbf S_{j-1} || \mathbf S_{j}$ (the concatenation of two sequences), and $\mathbf G_{j-1}^{m}$ with candidates survive steps (ii) and (iii).
\item[(v)] Increase $j$ by 1 and jump to step (i).  
\end{itemize}

\subsubsection{Error Tolerance} 
The built Mask R-CNN may introduce errors in subtitle silhouette extraction due to model imperfections and interference from contents in the recordings. Such errors may further lead to inaccurate spatiotemporal vectors, generating invalid or even no candidates. We consider three types of errors: (i) \textit{Type 1 (Substitution)}, which involves classifying silhouettes into inaccurate categories; (ii) \textit{Type 2 (Deletion),} where valid subtitles are missed; (iii) \textit{Type 3 (Insertion),} which extracts non-existed silhouettes. 

For Type 1 errors, we handle them by adjusting the criterion of determining the consistency between spatiotemporal vectors ($\mathbf V_o$ and $\mathbf V_c$) of the target silhouette sequence and a potential candidate video clip. Particularly, for step (2) of Spatiotemporal Correlation, as presented in Section~\ref{subsubsec:sub_demo}, we calculate the Euclidean distance $d$ between the two spatiotemporal vectors, i.e., the square root of the sum of the squared differences between the two vectors. We can then set up a threshold $d_0$, and if $d>d_0$, both vectors are regarded as inconsistent; otherwise, they will be regarded as consistent. Thus, we can control $d_0$ to adjust the tolerance against type 1 errors. Empirically, we enable $d_0^2= \lceil 0.1 \times |\mathbf V_o| \rceil$, always successively overcome substitution errors, where $|\mathbf V_o|$ denotes the total number of silhouettes, and $\lceil x \rceil$ denotes the ceil function, returning the smallest integer greater than or equal to $x$.

To handle Type 2 or 3 errors, we develop a heuristic solution by guessing and adding or deleting erroneous elements in the silhouette sequence. We assume there are  $D$ deletion errors and $I$ insertion errors. 
For deletion errors, we consider that each missing silhouette can be at any position in the sequence, and it can be divided into any cluster according to its height. When we add a missing silhouette into a silhouette sequence with length $m$, there are $m$+1 positions to choose from, and its mapping $l_{miss}$ in the spatiotemporal vector of the new silhouette sequence will match with any value, i.e., $|l_{miss}- l|  =0$, where $l$ denotes any value. Similarly, regarding insertion errors, we consider that the extra incorrectly recognized silhouette can be any element in the silhouette sequence. For each insertion error, we thus iteratively remove one element from the silhouette sequence, starting from the first and proceeding to the last. In general, $D$ and $I$ are bounded. We demodulate all resultant possible silhouette sequences and combine their returned results to form the set of final candidates. 

\subsubsection{Impact of User Operations.}
If the victim pauses the video and resumes later, a complete silhouette sequence can still be obtained, while the recording duration $T$ no longer reflects the actual duration of uninterrupted video playback. The attacker can detect pauses (e.g., by comparing pixel differences of successive frames) and get the pause interval ($T^{\prime}$). Next, the duration of the silhouette sequence will be updated as $T-T^{\prime}$, eliminating the impact of pausing.

When the victim rewinds or fast-forwards a video from time $t_1$ to $t_2$, the attacker would obtain two separate video clips with respective silhouette sequences, one before time $t_1$, and one after $t_2$. The inter-sequence duration $T_{1,2}=t_2-t_1$ no longer correctly reflects the duration between the two clips when the video plays normally. However, the candidates for the two silhouette sequences should belong to the same video.
Also, for rewinding and fast-forwarding, the starting time of the second clip should be smaller and larger than the ending time of the first clip, respectively. Thus, we need to revise the joint demodulation algorithm. We first demonstrate each silhouette sequence. With each candidate pair ($\mathbf G_1$, $\mathbf G_2$), where $\mathbf G_i$ ($i \in \{1,2\}$) is one candidate for the $i^{th}$ silhouette sequence, we check whether both $\mathbf G_1$ and $\mathbf G_2$ belong to the same video. If not, we discard this candidate pair; otherwise, we continue to check whether the starting time of $\mathbf G_2$ and the ending time of $\mathbf G_1$ satisfy the above requirement. If not, we discard the candidate pair, otherwise, this candidate pair will be then added to the final candidate set. 

\section{Evaluation} 
\label{sec:exp}

We develop an app to implement \textit{SilhouetteTell}, which can run on an off-the-shelf Android/iPhone smartphone. 

\subsection{Experimental Setup}

The attacker aims to identify a victim's video in a known set. This problem aligns with existing video identification studies~(e.g.,~\cite{RoeiBeauty,bae2022watching,LakshmananOn}). We construct the dataset by capturing 300 subtitle files from the three most popular media providers (YouTube, Amazon Prime Video, and Netflix). 
For each provider, we randomly select 100 movies. The duration of a selected movie ranges from 33 to 200 minutes, with the number of subtitles varying between 413 and 4,071.
{\color{black}This 300-video dataset is solely for evaluation.} The victim watches a video in the dataset, whether online or offline, on a typical personal device using the corresponding streaming service. We consider an attacker who is at a distance (4 meters or above) from the victim and uses their smartphone camera to record a video of the victim's screen. The recorded video is consistently set to 1080p (1920 $\times$ 1080 pixels) at 60 FPS, which is a common camera setting for today's phones. We also investigate the impact of the recording resolution in Appendix~\ref{subsubsec:resolution}. 

We calculate \textit{top-k accuracy} in terms of video title or clip, which is defined as the probability that the top $k$ guesses from the obtained $N$ candidates contain the target. If $k> N$, we have $\alpha =1$; otherwise, $\alpha=\frac{\Mycomb [1] {1} \cdot \Mycomb [(N-1)] {(k-1)}}{\Mycomb [N] {k}}= \frac{k}{N}$, where $\Mycomb [N] {k}$ is the number of combinations by choosing $k$ from $N$ numbers. 

Meanwhile, we also utilize traditional text/image recognition algorithms (Tesseract OCR and CRNN for text recognition, and ClipCap for image recognition) to test whether they can identify meaningful information from the recorded blurry videos.  

\begin{figure}[t]
\centering \includegraphics[width=2.5in]{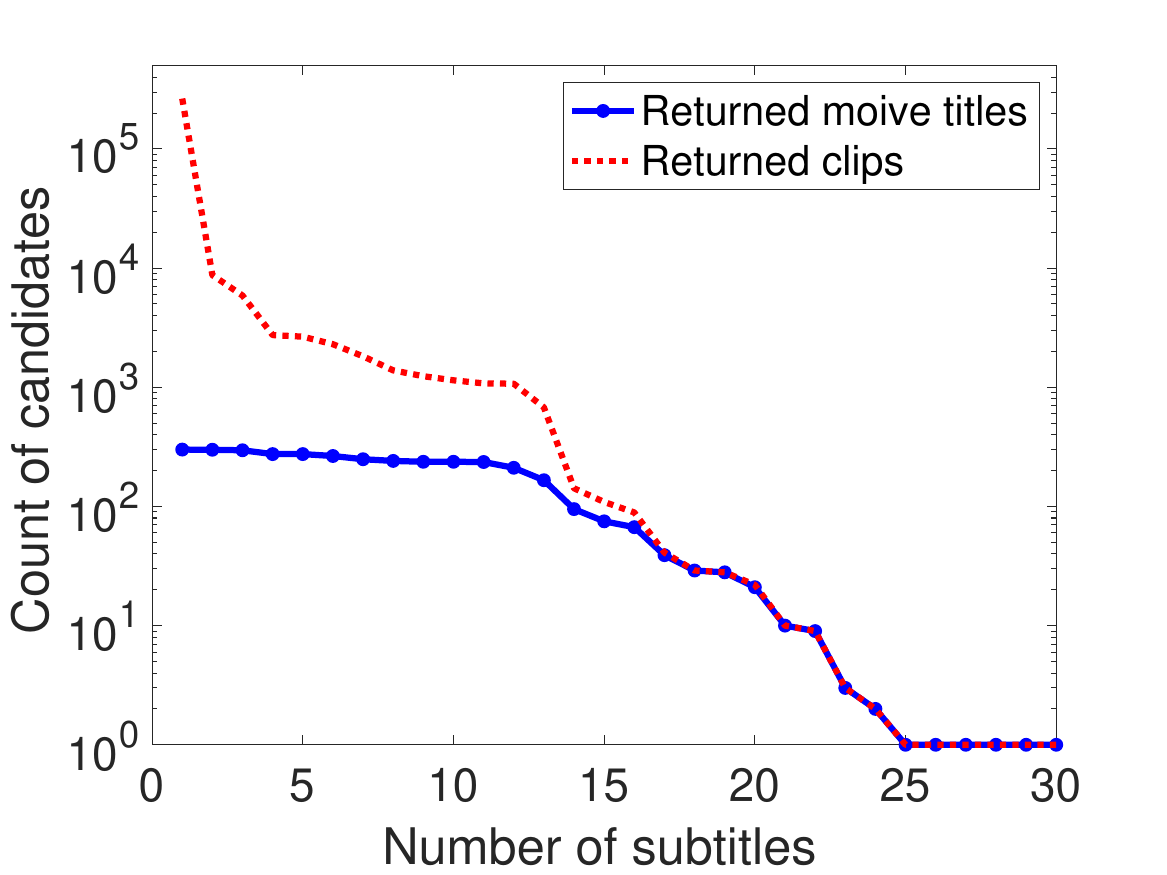}
\vspace{-0.1in} 
\caption{The evolution of the count of candidates.} 
\label{fig:number_cmp}
\vspace{-17pt}
\end{figure}

\subsection{Case Study}
\label{subsec:case}

In this example, the victim watches a movie (``The Tomorrow War") on Amazon Prime Video with a 16-inch MacBook Pro. The attacker uses a Samsung Galaxy Z Fold4 smartphone to record a 2-minute video targeting the victim's screen while the video plays from its $2,208^{th}$ to $2,327^{th}$ second. The attacker thus obtains a sequence of 30 subtitle silhouettes. The returned results from OCR are always nonsensical strings; the results of CRNN are empty; by enforcing ClipCap, the results are irrelevant to the scene of the frame. 

Figure~\ref{fig:number_cmp} presents the counts of returned video titles and clips using \textit{SilhouetteTell} during the processing of the silhouettes. For a video with $T_0$ seconds, we consider that it contains $N_c$ $T$-second video clips, with a second as the basic unit, i.e., $N_c=T_0-T+1$. We see for the initial 7 silhouettes, the number of returned video titles has almost no change, remaining equal to or slightly smaller than the total size (i.e., 300) of the suspect video library. This is because the spatiotemporal structure of a short sequence with fewer silhouettes is common, leading to the failure of using it to distinguish varying videos. On the clip level, as different clips with varying subtitles have different periods, the number of returned video clips decreases accordingly.   
Also, with the number of silhouettes increasing, the numbers of matching video titles and clips both gradually converge to 1, i.e., the correct match, indicating successful recognition. 

\subsection{Robustness to Influential Factors}
\label{subsec:factors} 
The recording duration/distance/angle/device may vary for the attacker; the victim may use different watching devices and playback speeds. We evaluate their impact. For recorded videos, we also use traditional recognition methods (OCR, CRNN, and ClipCap) to extract meaningful information, and none achieves video inference. The impact of the playback speed is presented in Appendix~\ref{subsubsec:speed}. 

\begin{figure}[t]
\centering \includegraphics[width=3in]{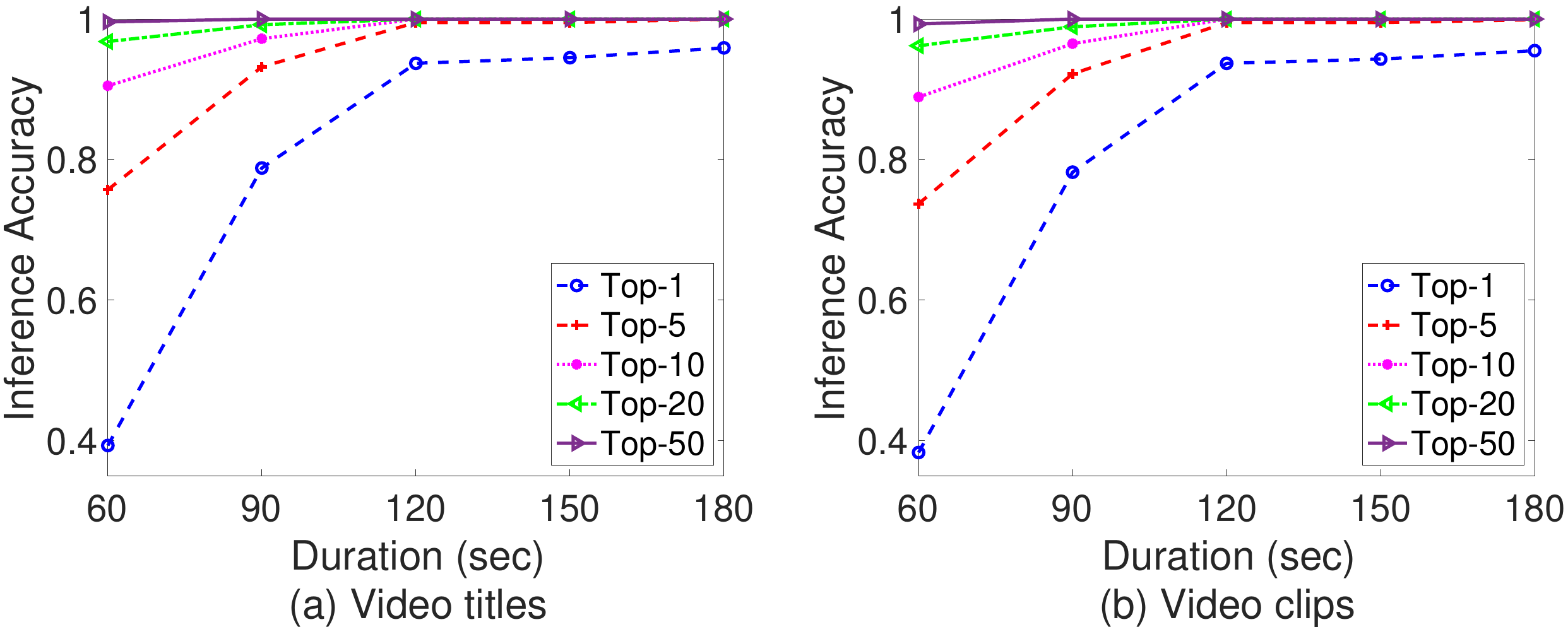}
\vspace{-0.1in} 
\caption{Average top-$k$ accuracy vs. recording duration.} 
\label{fig:duration}
%\vspace{-0.1in}
\end{figure}
\subsubsection{Impact of Recording Duration} 
\label{subsubsec:clip_duration} 
We vary the duration of the recorded video clip from 1 to 3 minutes, in increments of 30 seconds. For each clip duration, we randomly record 100 clips.  
Figures~\ref{fig:duration}a and~\ref{fig:duration}b present the inference performance in terms of the video title and clip. We see from 60 to 120 seconds, the inference accuracy proportionally increases with time, and it maintains high when the duration is 120 seconds or above. When the recording period reaches 2 minutes, the top-10 accuracy stays 100\%, and the top-1 accuracy is above 93.7\%. Some selected video clips have fewer subtitles due to scenes containing no subtitles (such as gunfight scenes). The spatiotemporal feature of such a video clip may not be rich enough to make the clip distinct, generating multiple candidates. To achieve a desirable inference performance, we employ 2 minutes as the recording period for the following discussion. 

\begin{figure}[t]
\centering \includegraphics[width=3in]{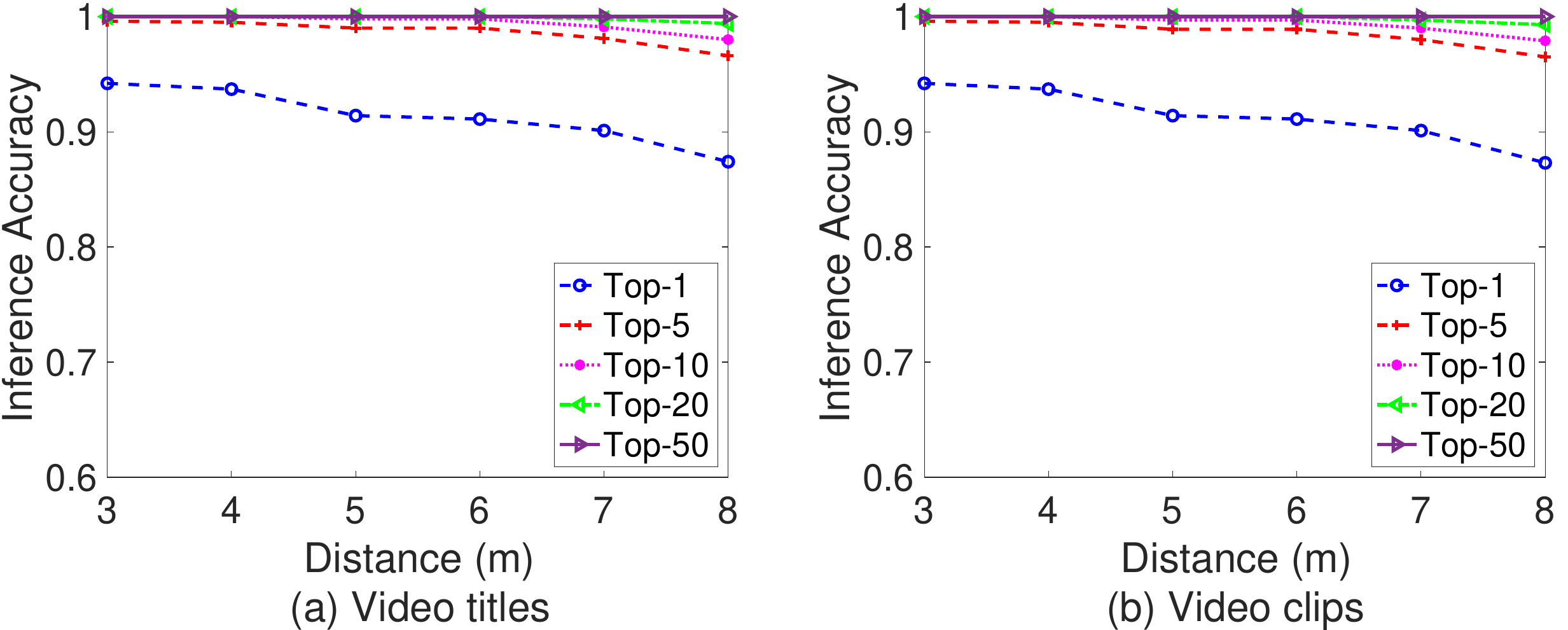} 
\vspace{-0.1in} 
\caption{Average top-$k$ accuracy vs. recording distance.} 
\label{fig:recording_distance}
\vspace{-0.15in}
\end{figure}
\subsubsection{Impact of Recording Distance}
\label{subsubsec:distance} 
A traditional shoulder surfer must stay close (e.g., within 2 m) to the victim to directly observe their screen, but this proximity is likely to raise suspicion~\cite{ye2017cracking}. To avoid suspicion, we vary the recording distance from 3 to 8 m, in increments of 1. For each recording distance, we randomly record 100 clips.
Figure~\ref{fig:recording_distance} shows the inference performance for different recording distances. We see the inference accuracy remains high for varying distances. Also, with the distance increasing, the average top-1, top-5, and top-10 accuracy values for both video titles and clips slightly decrease. When the distance is 7 m, the average top-1 accuracy values for video titles and clips both exceed  90\%. 
Besides, the top-20 and top-50 accuracy values are always 100\%.

\begin{figure}[t]
\centering \includegraphics[width=3in]{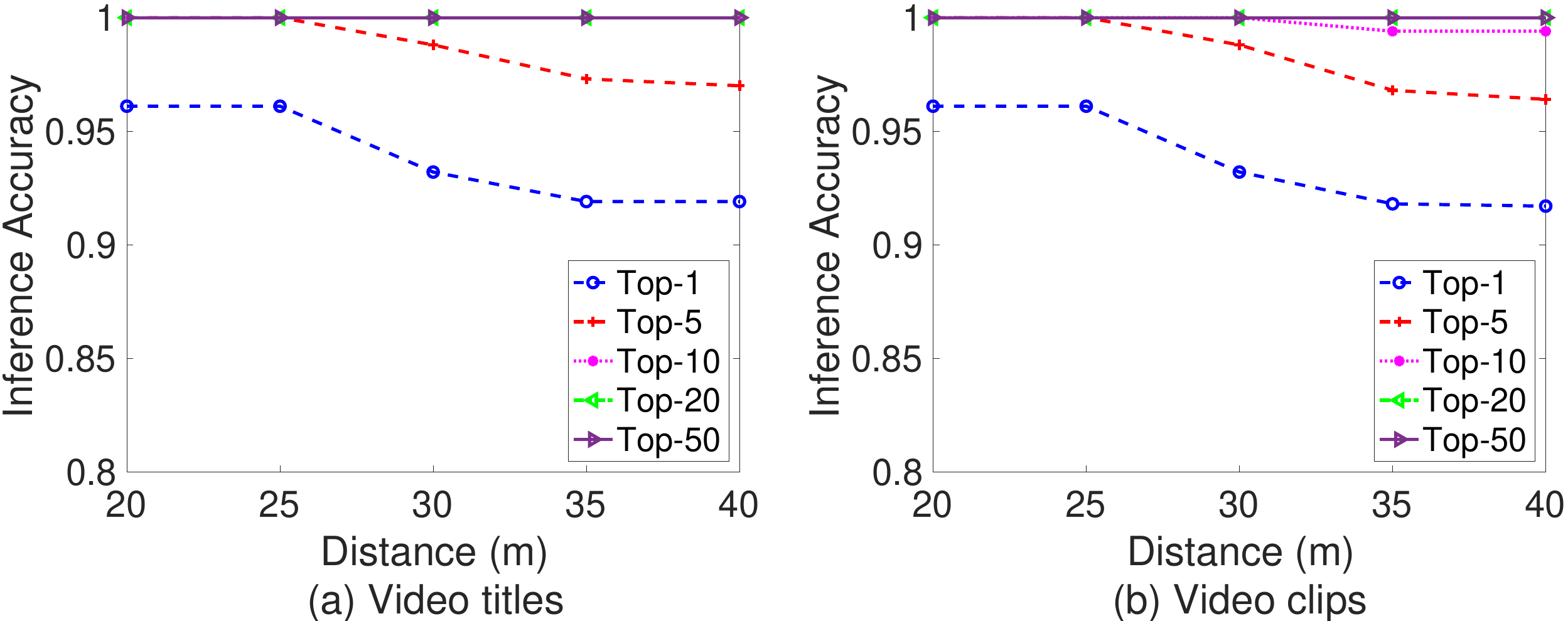} 
\vspace{-0.1in} 
\caption{Inference accuracy in long recording distance tests.} 
\label{fig:long_d}
%\vspace{-0.05in}
\end{figure}

\textbf{Long Recording Distance Tests: } 
Unlike digital zoom, which merely enlarges pixels without enhancing resolution, some smartphones feature optical zoom lenses that adjust the optics to bring the target closer while preserving image clarity. We employ a Samsung S23 Ultra smartphone with 10 times optical zoom~\cite{GalaxyS23}. Our long-distance tests were conducted on a 40 m long corridor, as shown in Appendix~\ref{appendix:long-distance}. 
%Figure~\ref{fig:long_distance}. 
We vary the recording distance from 20 to 40 m, in increments of 5, and perform 30 trials for each distance. Figure~\ref{fig:long_d} shows the obtained average top-$k$ accuracy. We see that the top-1 accuracy values for video titles and clips slightly decrease with the recording distance but still exceed 92\% even at a distance of 40 m. For both video titles and clips, the top-10, top-20, and top-50 accuracies stay at 100\% across all distances; the top-5 accuracy is near 100\% when the distance ranges from 20 to 30 m and slightly drops to about 97\% when the distance equals 35 or 40 m.

\subsubsection{Impact of Recording Angle}
\label{subsubsec:angle} 

\begin{figure}[t]
\centering \includegraphics[width=3.3in]{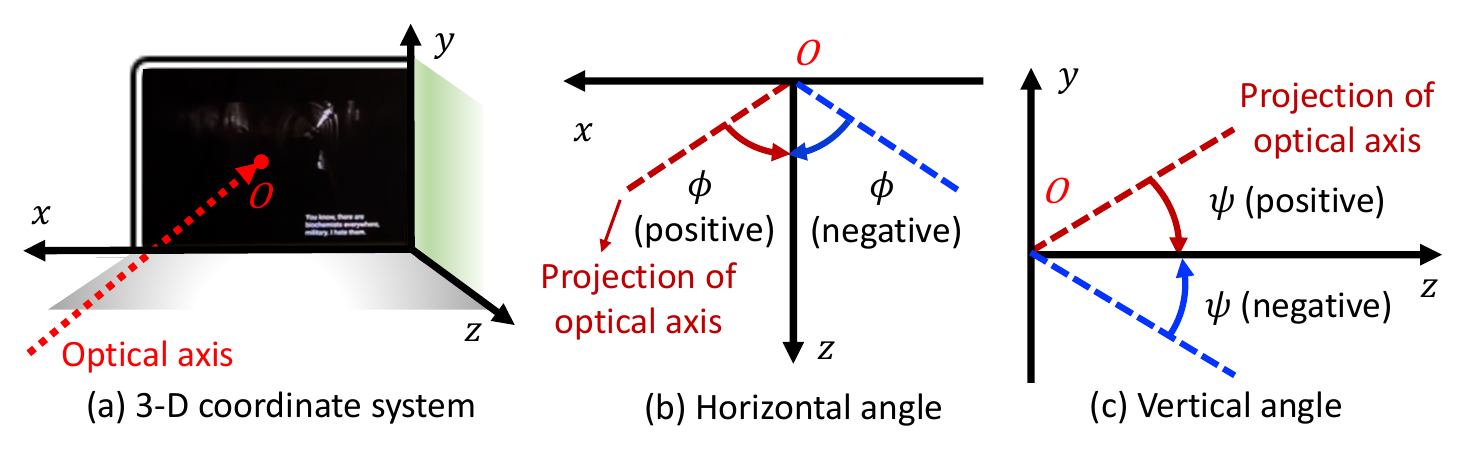} 
\vspace{-0.1in} 
\caption{Horizontal and vertical angles.} 
\label{fig:angles}
\vspace{-0.15in}
\end{figure}

\begin{figure*}
\begin{minipage}[t]{0.32\linewidth}
\centering
\includegraphics[width=2.2in]{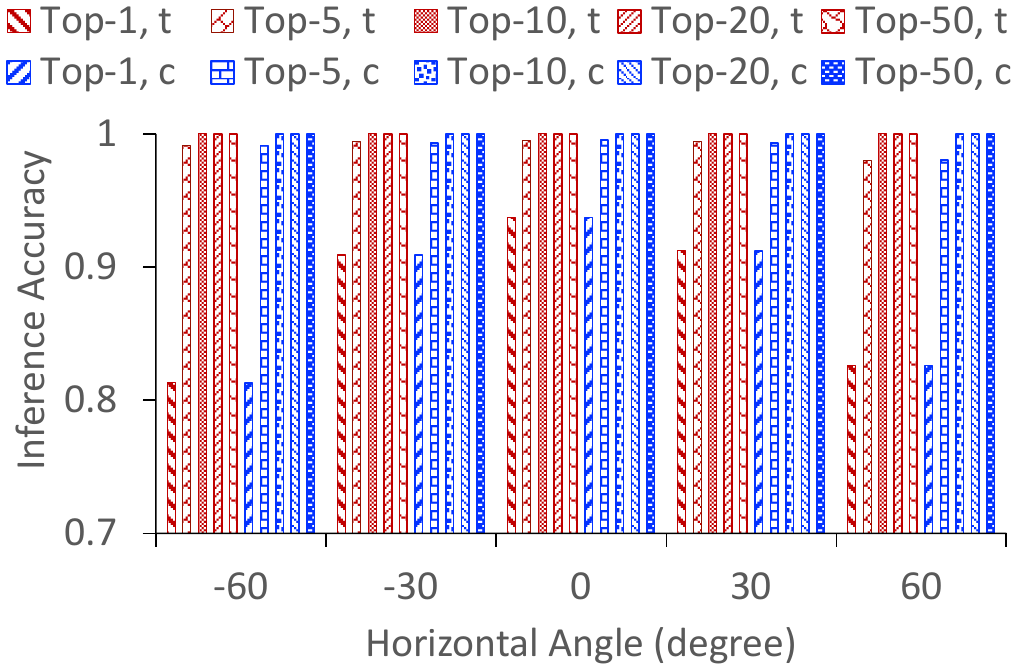}  
\vspace{-0.12in}
\caption{Impact of horizontal angles.} 
\label{fig:impact_hor}
\end{minipage}
\hspace{0.03in}
\begin{minipage}[t]{0.32\linewidth}
\centering
\includegraphics[width=2.2in]{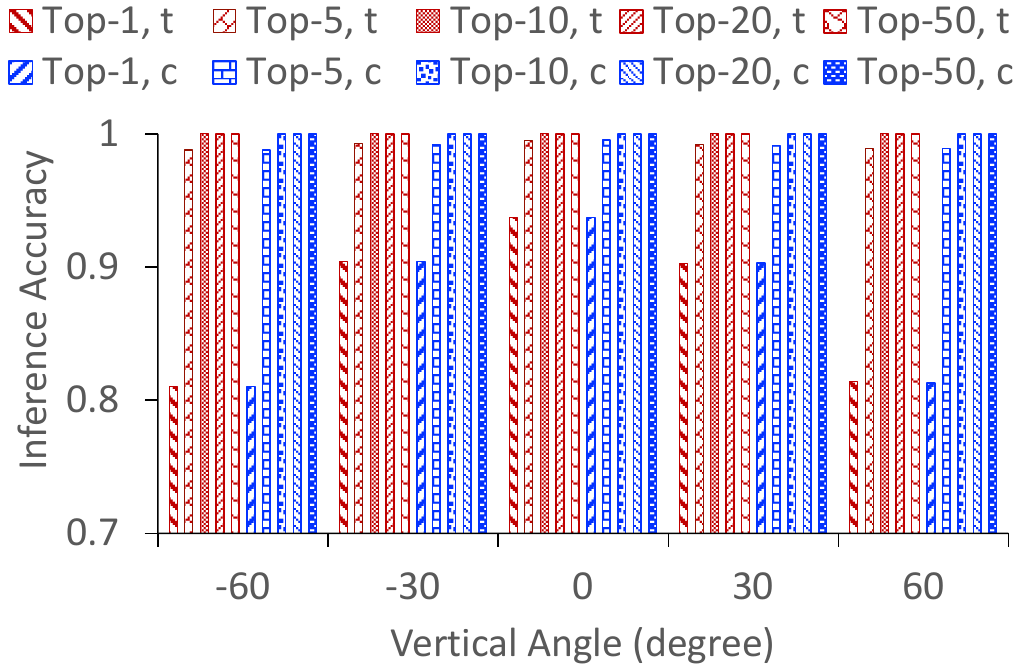} 
\vspace{-0.12in}
\caption{Impact of vertical angles.} 
\label{fig:impact_ver}
\end{minipage}
\hspace{0.03in}
\begin{minipage}[t]{0.32\linewidth}
\centering
\includegraphics[width=2.2in]{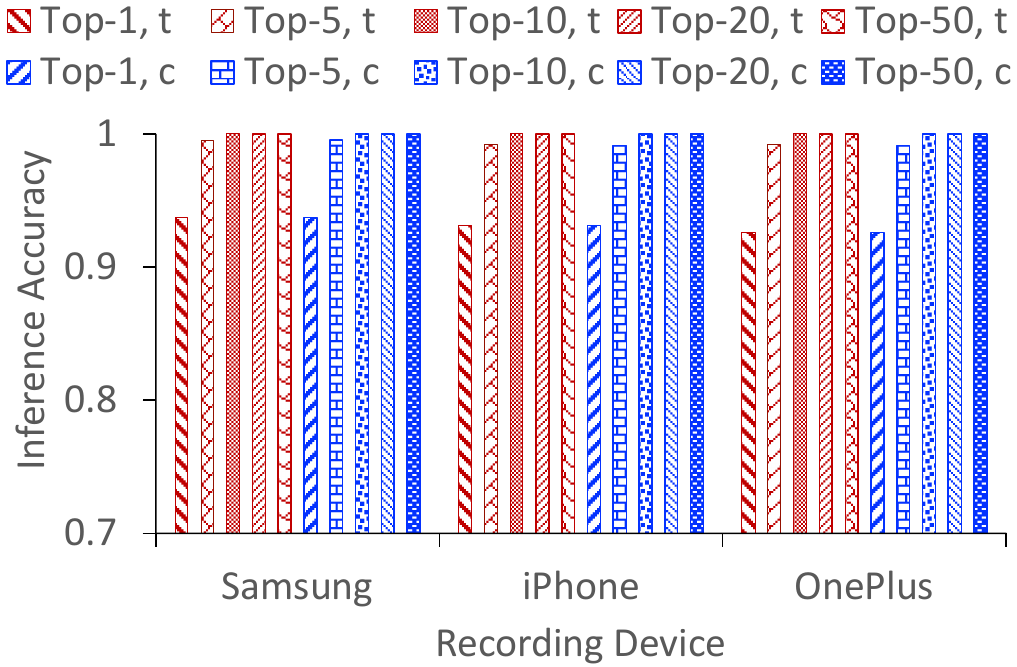}
\vspace{-0.12in}
\caption{Impact of recording devices.} 
\label{fig:record_d}
\end{minipage}
\vspace{-0.1in}
\end{figure*}

As shown in Figure~\ref{fig:angles}, the optical axis of the camera intersects the screen at point $O$; the angle between the projection of the optical axis in the $xz$-plane and the vertical line (i.e., $z$ axis), which passes through the point $O$ and also is perpendicular to the screen, is referred as horizontal angle ($\phi$); the angle between the projection of the optical axis in the $yz$-plane and the vertical line is referred as vertical angle ($\psi$). We vary $\phi$ and $\psi$ both from -60$\degree$ to 60$\degree$, in increments of 30$\degree$. We maintain one of two angles fixed and perform 100 trials of \textit{SilhouetteTell} for each value of the other angle, with each trial randomly recording a video clip. 
Figures~\ref{fig:impact_hor} and \ref{fig:impact_ver} present the corresponding average top-$k$ accuracy. ``Top-$k$, t'' and ``Top-$k$, c'' refer to top-$k$ accuracy for video titles and clips, respectively.
We see the top-$k$ accuracy values for video titles and clips are consistently high. This appears as there is usually one video clip in a video emerging as the candidate. Also, as the recording angle ($\phi$ or $\psi$) increases, the top-1 and top-5 accuracies for video titles and clips all slightly decrease. This is because a larger recording angle may result in more distortion of subtitle silhouettes, making the spatiotemporal features of the extracted silhouette sequence less distinguishable and thus producing more candidates. When $|\phi| \!\leq\!30\degree$ or $|\psi| \!\leq\!30\degree$, \textit{SilhouetteTell} achieves a top-1 accuracy exceeding 91\% for either video titles or clips. Besides, the top-10, top-20, and top-50 accuracy values for all cases are 100\%. 

\subsubsection{Impact of Recording Device}
\label{subsubsec:recording_device} 

We experiment with three popular smartphones, Samsung Galaxy Z Fold4, iPhone 15 Pro Max, and OnePlus 12 (referred to as Samsung, iPhone, and OnePlus, respectively), and use each to randomly record 100 video clips. Other factors (e.g., recording distance, angle, and resolution) maintain the same. Figure~\ref{fig:record_d} shows the obtained inference accuracy. We see that top-$k$ accuracy values of all recording devices for video titles and clips are consistently high, and there is no obvious difference in attack performance among varying recording devices.

\subsubsection{Impact of Watching Device}
\label{subsubsec:watching_device} 

\begin{figure*}
\begin{minipage}[t]{0.32\linewidth}
\centering
\includegraphics[width=2in]{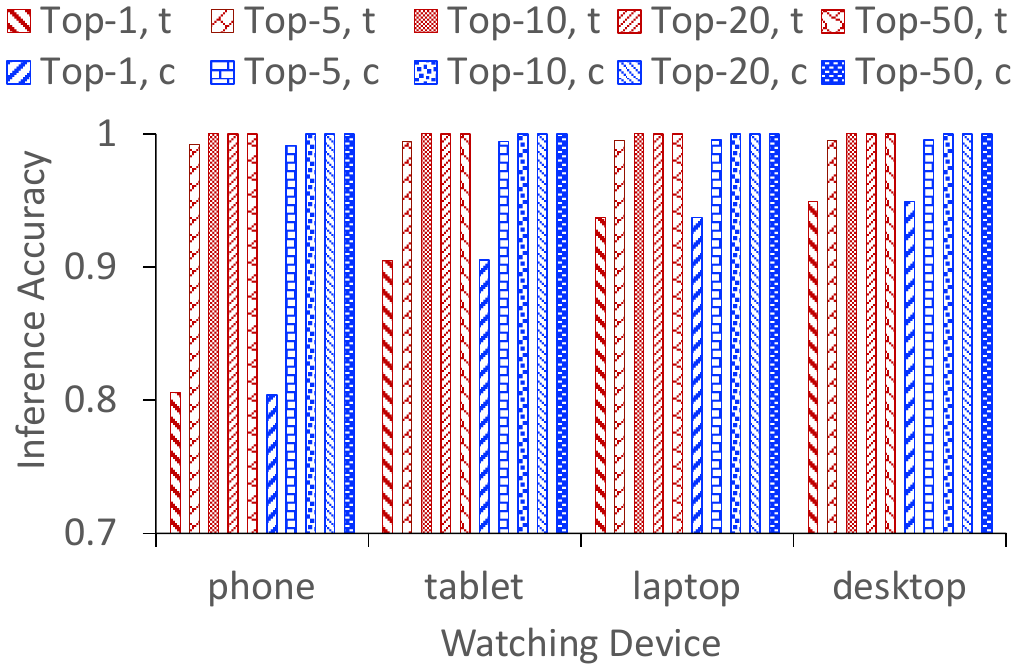} 
\vspace{-0.12in}
\caption{Impact of watching devices.}
\label{fig:watch_d}
\end{minipage}
\begin{minipage}[t]{0.32\linewidth}
\centering
\includegraphics[width=1.9in]{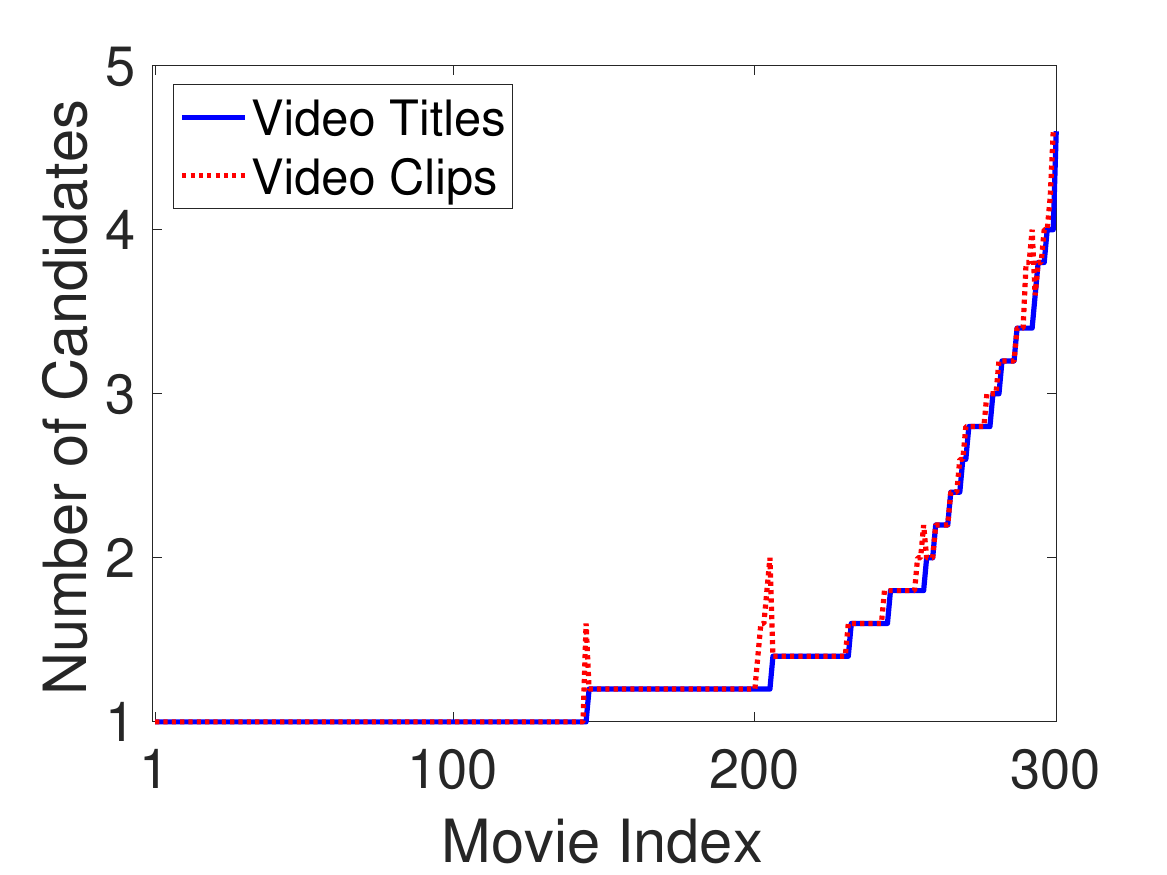}
\vspace{-0.12in}
\caption{Candidate count distribution.} 
\label{fig:overall_candidates}
\end{minipage}
\hspace{0.01in}
\begin{minipage}[t]{0.31\linewidth}
\centering
\includegraphics[width=2.1in]{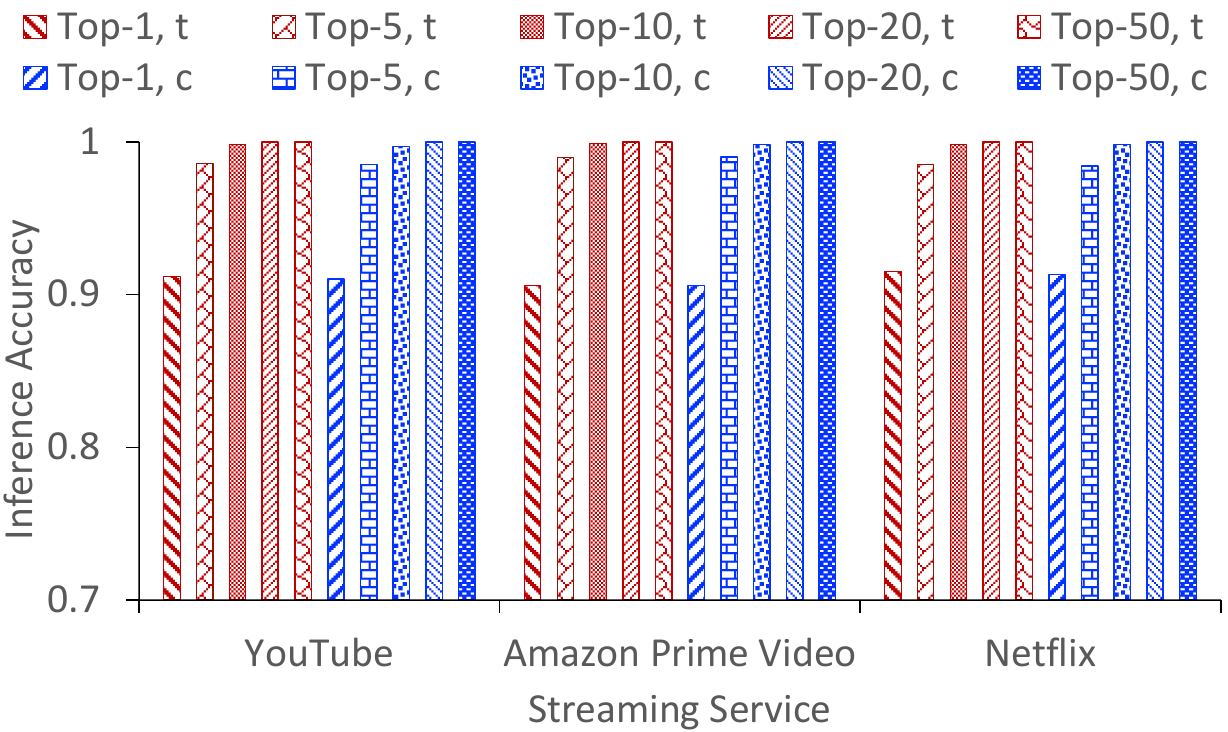}
\vspace{-0.12in}
\caption{Performance vs. services.} 
\label{fig:service}
\end{minipage}
\vspace{-0.05in}
\end{figure*}

A victim may watch videos on different devices, with varying screen and subtitle sizes. We test four common watching devices, a mobile phone (Samsung S23 Ultra) with a 6.8-inch screen, a tablet (iPad Pro) with a 12.9-inch screen, a laptop (MacBook Pro) with a 16-inch screen, and a desktop with a 23.8-inch ACER monitor. For each device, we randomly record 100 video clips. During recordings, we control all the rest factors consistently. Figure~\ref{fig:watch_d} plots the resultant inference performance. We see that for both video titles and clips, the top-1 and top-5 accuracy values increase with the screen size, while the top-10, top-20, and top-50 accuracy values maintain 1. Particularly, when the victim watches videos with a smartphone, our attack still achieves top-1 accuracy of above 81\% for inferring video titles or clips.

\subsection{Overall Attack Impact}
\label{subsec:overall} % of the suspect video library, in which YouTube, Amazon Prime Videos, and Netflix provide 100 movies, respectively
We perform 5 different attack trials for watching each video in the built dataset, and thus have $5 \!\times\! 3 \!\times \!100 \!=\! 1,500$ video clips in total. 
For comparison, we also decode these blurry video clips with OCR, CRNN, and ClipCap, and fail to extract any useful information. % for inferring video titles or clips.  

We sort the movies in ascending order of the candidate video title number, and for movies whose candidate video titles are the same, we further sort them in ascending order of the candidate video clip number. We then index the sorted movies from 1 to 300 in increments of 1.
Figure~\ref{fig:overall_candidates} presents the corresponding average numbers of video title and clip candidates when the victim watches each movie. We see that the average video title and clip candidate counts are consistently low (under 4.6). Particularly, there are 143 movies whose video title and clip candidate counts are both 1, indicating that our attack can uniquely identify the video title and clip. Figure~\ref{fig:service} shows the average top-$k$ accuracy across different streaming services. We see that all top-$k$ accuracy ($k\in\{1,5,10,20,50\}$) values for both video titles and clips are consistently high (above 91\%). Also, there is no obvious performance difference for varying streaming services. These results convincingly indicate that our attack is robust against different movies and service providers. In addition, we also examines the effects of video libraries of different sizes, as presented in Appendix~\ref{appendix:scalability_analysis}. 

\textbf{Side-Channel Entropy Analysis:}
We use a subtitle silhouette sequence within a 2-minute sliding time window as the analysis unit to evaluate the entropy of the proposed subtitle silhouette side channel.
Let $T_v$ denote the total number of distinct subtitle sequences. The corresponding $T_v$ subtitle silhouette sequences result in $T_v$ spatiotemporal vectors. By clustering the same vectors into the same class, we obtain $\mathcal{U}$ unique classes. %We use \textbf{entropy} to measure the 
Consequently, the subtitle silhouette sequence $X$’s entropy $H(X)$ can be computed as $-\sum^{\mathcal{U}}_{i=1} P(x_i) log_2{P(x_i)}$, where $P(x_i)$ represents the probability that $X=x_i$ holds. 
From our dataset of 300 videos, we extract 731,228 unique classes of spatiotemporal vectors. The largest class contains 14,370 elements, while the smallest class contains only a single element. Accordingly, the maximum entropy is $\log_2{(14,370)} \approx 13.81$ bits, and the minimum entropy is 0 bits. Also, the average entropy equals $2.58\times 10^{-5}$ bits, indicating low uncertainty and a highly predictable structure in the subtitle silhouette side channel. 

%\vspace{-0.2in}
\textbf{Open-World Scenario:} 
We also investigate the performance of \textit{SilhouetteTell} in an open-world setting, where the target video is not present in the fingerprint database. Specifically, we randomly divide the 300 videos into 10 groups, each containing 30 videos. We then iteratively use each group as the fingerprinting target set, while the remaining nine groups serve as the open-world database. For each target video, we randomly sample 10 clips at a given recording duration, resulting in a total of $10\times 30 \times 10 = 3,000$ clips per duration. We consider five different recording durations, including 60, 90, 120, 150, and 180 seconds.  

\begin{table}[t]
\centering
%\vspace{-0.15in}
\caption{False positive rates by duration in open-world setting (``Avg." denotes average; ``\#" represents ``number").} 
\vspace{-0.15in}
\label{tab:openworld_fp}
{\begin{tabular}{lccccc}
\toprule
\textbf{\!\!\!\makecell{Duration \\ (sec)}\!\!} &
\textbf{\!\!\makecell{\# of FP \\ Clips}\!\!} & \!\!\textbf{\makecell{FP \\Rate (\%)}} \!\!&
\!\!\!\!\textbf{\makecell{Avg. \# of \\Matched Clips}}\!\!\!&\!\!\! \textbf{\makecell{Avg. \# of \\ Matched Titles}} \!\!\!\!\\
\midrule
60   & 312 & 10.4 &  7.33 & 3.85 \\
90   & 216 & 7.2  &  5.27 & 2.30 \\
120  & 109 & 3.6  &   1.39 &  1.38 \\
150  &  40 & 1.3  &    1.30 &  1.15 \\
180  &  29 & 1.0  &    1.24 &  1.10 \\
\bottomrule
\end{tabular}}
\vspace{-0.15in}
\end{table}

For each target clip, we find that the inference result is either empty or consists of a list of clips from the open-world database. A clip is considered a false positive if it matches any clip in the open-world database. Table~\ref{tab:openworld_fp} presents the false positive rates and the average matched clips and titles for false positives in the open-world setting. We observe that with the recording duration increasing, the FP rate significantly decreases. For a recording duration of 60 seconds, 312 out of 3,000 clips are incorrectly matched, resulting in a false positive (FP) rate of 10.4\%. As the recording duration increases to 120 and 180 seconds, the FP rate decreases to 3.6\% and 1.0\%, respectively. 
In addition, the average number of matched clips and titles per false positive case also decreases with longer recording durations. These results indicate that increasing the recording duration reduces the ambiguity in spatiotemporal patterns of subtitle silhouettes, leading to lower FP rates. 

In the closed world, \textit{SilhouetteTell} achieves 90.9\% top-1 accuracy for clip recognition, indicating that for most library videos, \textit{SilhouetteTell} can correctly recognize it, even when provided with only a single candidate. Meanwhile, in the open world, with a recording duration of \(120\)\,s, \textit{SilhouetteTell} outputs a non-empty candidate list for only 109 out of 3,000 clips that are not in the library, and correctly determines that the remaining are absent from the library.

\begin{figure}[t]
\centering \includegraphics[width=3in]{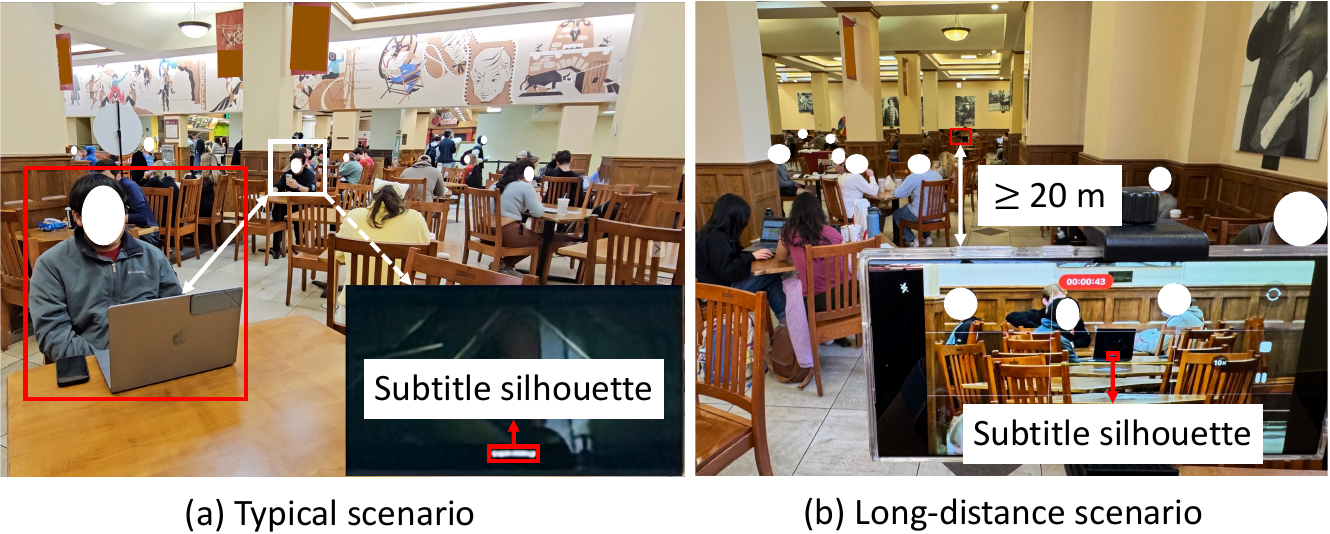} 
\vspace{-0.1in} 
\caption{Real-world scenarios.} 
\label{fig:real}
\vspace{-0.15in}
\end{figure}

\subsection{Controlled Experiments}
%}
\label{subsec:user_study}

We recruited 10 volunteers (U1-U10; aged 23-32 years old; 5 females and 5 males).\footnote{This study has been reviewed and approved by our institution’s IRB.}
We perform experiments in a cafeteria-like environment, where people may walk around or move chairs. Each participant was instructed to sit at a table and watch a movie from the suspect library on a 16-inch MacBook Pro laptop. Two typical cases are considered, as shown in Figure~\ref{fig:real}: (a) a general scenario where there is no need to have the optical zoom and adjust it to bring the target closer and we run \textit{SilhouetteTell} on a Samsung Galaxy Z Fold4 smartphone at a distance above 4 m behind the victim; (b) a long-distance scenario, where we run \textit{SilhouetteTell} on a Samsung S23 Ultra smartphone and turn on 10 times optical zoom towards the target at a distance of above 20 m away from the victim, making the attack more difficult to be noticed by the victim.

\begin{figure}[t]
\centering \includegraphics[width=3.3in]{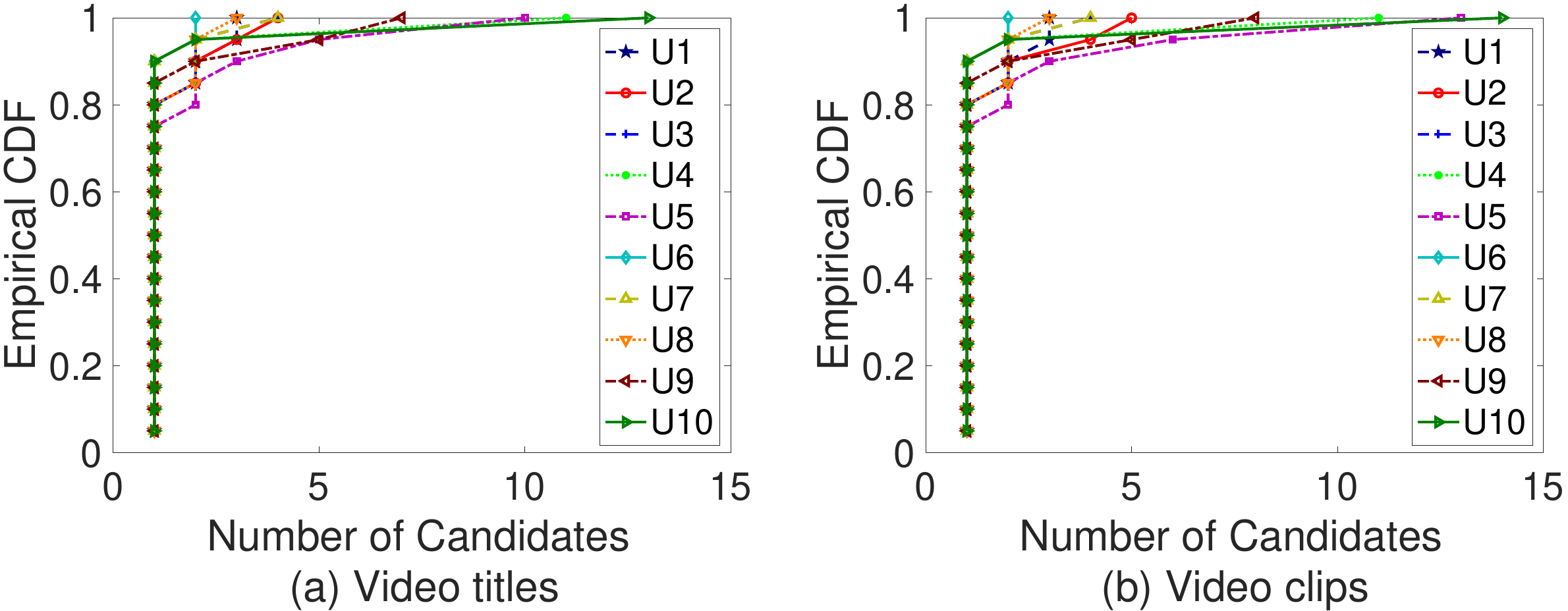}
\vspace{-0.1in} 
\caption{CDFs of the numbers of obtained candidates.} 
\label{fig:user_cdf}
\vspace{-0.05in} 
\end{figure}

\begin{figure}[t]
\centering \includegraphics[width=3.3in]{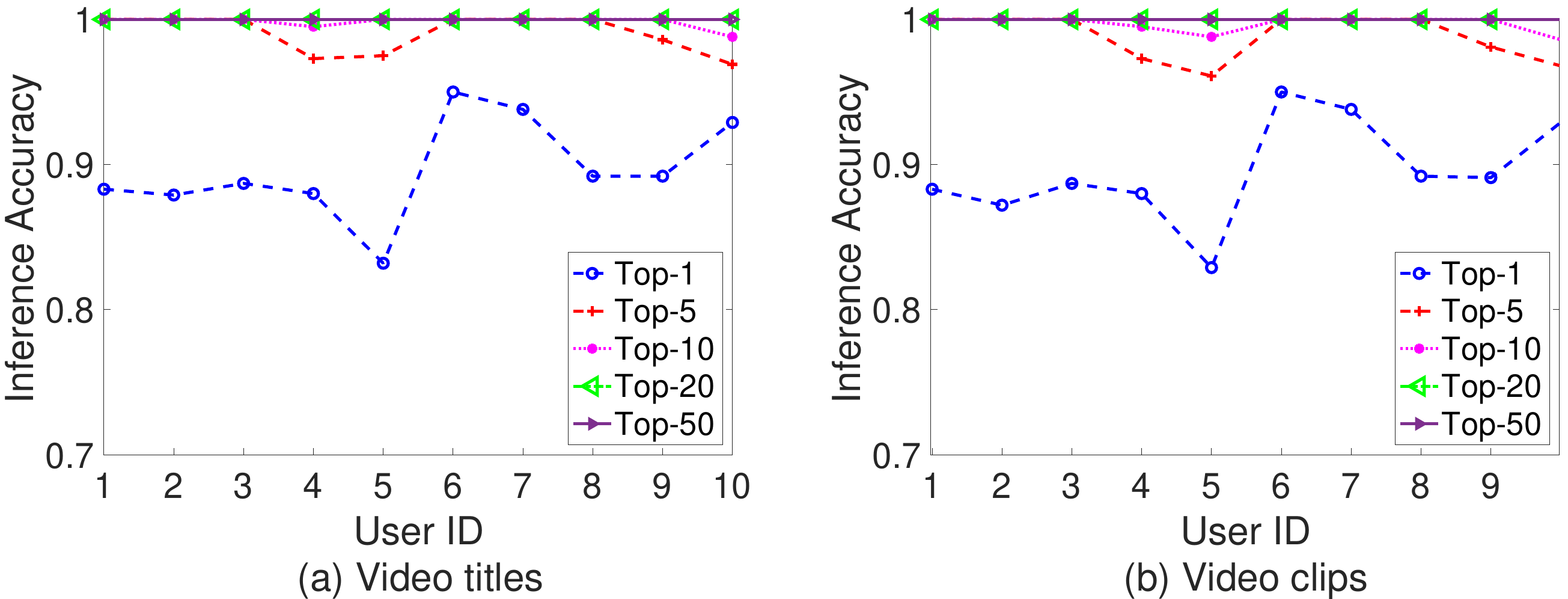}
\vspace{-0.1in} 
\caption{Average top-$k$ accuracy in typical scenario.} 
\label{fig:top_user}
\vspace{-0.25in}
\end{figure}

Participants can freely adjust their sitting positions and choose to pause, fast-forward, or rewind videos as usual. For each participant, we performed 20 independent attempts with random video clips selected by the participant. Similarly, for comparison, we utilize OCR, CRNN, and ClipCap to process each recording and generate no meaningful information related to the target video. We present the results of \textit{SilhouetteTell} to the participant, who determines whether the watched clip is in the inferred list. For all trials, the target clip is always in the inferred result. Figure~\ref{fig:user_cdf} plots the CDFs of the numbers of obtained candidates per user for video titles and clips. We see that the maximum numbers of candidates among all users for video titles and clips are 13 and 14, respectively. Also, regardless of users, the probabilities of uniquely inferring the target video title and clip are both at least 75\%. Figure~\ref{fig:top_user} presents the inference performance. We observe that our attack consistently achieves high average top-$1$ and top-$5$ accuracies (above 83\% and 97\% respectively) for all users in terms of video titles and clips; the average top-$10$, top-$20$, and top-$50$ accuracies are all near or equal to 100\%. 

We have similar observations in the long-distance scenario: for all users, the numbers of candidate video titles and clips range from 1 to 13, and from 1 to 15, respectively; the minimum average top-$1$ accuracies for video titles and clips are 88.9\% and 88.6\%, respectively. 

\textbf{Comparison with Baselines:}
For comparison, we evaluate the performance of human attackers using the two baseline methods: 
(i) Bare-eye recognition: 10 participants with normal or corrected-to-normal vision (either unaided or with glasses) attempt to recognize the videos from a distance of 40 meters, relying solely on their vision.  
(ii) 10$\times$ zoom recognition: The same participants repeat the above experiment using the 10$\times$ zoom lens of a Samsung Galaxy S23 Ultra phone. 
In each trial, we randomly select a video from the 300-movie dataset and play a 2-minute clip. Each participant performs 30 attempts per above baseline. The results show that no participant correctly identifies any movie under bare-eye or 10$\times$ zoom conditions, yielding a 0\% success rate for video identification.

In addition, we also test a visual input attack, i.e., CNN-LSTM based recognition. The CNN-LSTM hybrid architecture~\cite{donahue2015long,xu2015show} is a well-established model for spatiotemporal visual understanding tasks. We use the ResNet-50 network as the backbone of the spatial feature extractor. The extracted spatial features are fed into an LSTM layer to capture temporal dependencies. We fine-tune the LSTM and final classification layers using our dataset of 300 videos. 

To generate the dataset for building the model, we adopt a grid-based multi-angle recording strategy. We vary the horizontal viewing angle across five settings (0°, ±30°, ±60°), and the vertical angle similarly across five settings (0°, ±30°, ±60°), resulting in 25 distinct camera viewpoints. Also, we vary the recording distance from the target screen between 2 and 5 meters, in 1-meter increments. At each spatial configuration (i.e., a unique combination of angle and distance), we record each video once, yielding 100 recordings per video. We then get $300 \!\times\! 100\! = \!30,000$ recordings in total. We experiment with varying the number of recordings per video. To increase the dataset size, we randomly select 25 or 50 spatial configurations and add one new recording at each, resulting in 25 or 50 additional recordings per video. To reduce the dataset size, we similarly select 25 or 50 configurations at random and remove the existing recordings at those locations. This allows us to evaluate performance under different dataset sizes of 50, 75, 125, and 150 recordings per video. 
For each dataset size, we apply the standard 80:20 train-test split to the data of each video, ensuring that the CNN-LSTM model is trained with samples from all 300 videos (i.e., classes).

\begin{figure}[t]
\centering \includegraphics[width=2.3in]{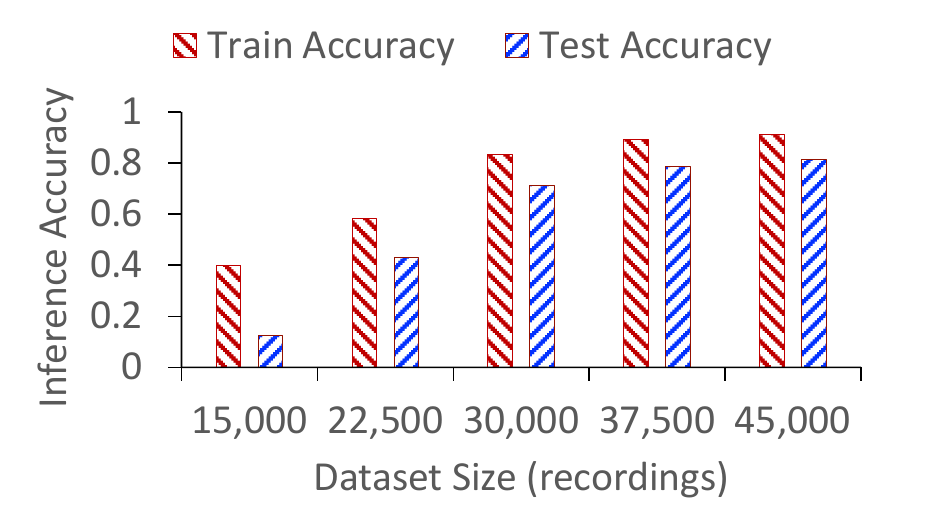}
\vspace{-0.23in} 
\caption{Performance under different dataset sizes.}  
\label{fig:accu_size}
\vspace{-0.22in}
\end{figure}

Figure~\ref{fig:accu_size} presents the obtained train and test accuracies of the CNN-LSTM model. We see that both train and test accuracies improve as the dataset size increases. With 100 recordings per class, the model achieves 83.3\% and 71.0\% accuracy for training and testing, respectively; increasing to 150 recordings per class raises these to 91.0\% and 81.0\%. To evaluate the model’s performance for longer recording distances, we record 20 different videos at distances of 6, 7, and 8 meters, respectively. The results show that inference accuracy declines as recording distance increases. The top-1 accuracies at 6, 7, and 8 meters are 0.60, 0.35, and 0.25, respectively. This decline is likely due to reduced spatial resolution and increased noise at greater distances, which degrade the model’s ability to extract meaningful features. These results demonstrate that the CNN-LSTM model, when trained with sufficient data, achieves high accuracy on configurations seen during training. However, its performance drops significantly under unseen recording conditions. Moreover, the data collection and training process is costly, which further limits the model's scalability. 

For clip-level recognition, \textit{SilhouetteTell} can achieve a top-1 accuracy of 0.91, while when 150 recordings per class are used for training, the top-1 accuracy for testing of CNN-LSTM attains is just 0.81. Meanwhile, with the recording distance increasing, the CNN-LSTM degrades sharply (top-1 $=0.60/0.35/0.25$ at $6/7/8$\,meters). In contrast, \textit{SilhouetteTell} shows only a slight decline, achieving top-1 accuracies of $0.91$, $0.90$, and $0.87$, respectively.

% These results show that \textit{SilhouetteTell} preserves high accuracy and stable performance at longer ranges, substantially outperforming the CNN-LSTM under the same condition.
\vspace{-0.07in}
\section{Discussions} 
\label{sec:discussion} 

\subsection{Limitation}
\label{subsec:limitation}

\hspace*{1em} \textbf{Necessity of Turning on Subtitles: }\textit{SilhouetteTell} does not work for cases when the victim watches a video without subtitles enabled. A recent survey shows that 50\% of Americans, and particularly 70\% of young viewers aged 18 to 25, watch content with subtitles most of the time~\cite{Surveysubtitle}. Also, subtitles are almost a prerequisite for following videos with foreign language content or in a noisy environment~\cite{VITACsubtitle}. Our attack thus poses a serious practical privacy threat.

\textbf{Unknown Subtitle Files: } \textit{SilhouetteTell} builds a suspect video library by downloading subtitles files and can only infer videos in the built library. Our experimental scale is comparable to that in most recent studies~(e.g.,~\cite{GuWalls,bae2022watching,ZhangTraffic}) but still limited. If the victim watches a subtitled video that does not appear in the library, the inference may fail. However, conducting larger-scale experiments by pre-downloading or pre-extracting subtitle files of more videos may further evidence the efficacy of \textit{SilhouetteTell}. 
\vspace{-0.1in}
\subsection{Defense Strategies}
\label{subsec:defense}

{\color{black} \subsubsection{Disabling Subtitles} Intuitively, to}
thwart \textit{SilhouetteTell}, a user may close subtitles, which, however, are necessary in many cases, e.g., for translating foreign-language content. Also, disabling subtitles may reduce viewing experience. 

{\color{black} We randomly select 25 2-minute video clips and play each with subtitles off. As no subtitle silhouettes appear in the recorded video,  \textit{SilhouetteTell} fails to recognize the video title/clip across all trials.}

 {\color{black}\subsubsection{Rendering Obfuscated Subtitle Silhouettes} 
We may confuse the attacker by rendering} the same or randomized subtitle silhouettes. Currently, the shapes (e.g., line counts) of a video's subtitles display on the screen strictly following the video's subtitle file. We can pad each subtitle of a video frame with special characters, which aims to prevent disclosing the original true subtitle silhouettes. As a result, there is no confirmed correlation between a video clip's subtitles and the observed corresponding subtitle silhouettes. \textit{SilhouetteTell} would thus fail. This method, however, may increase the complexity of rendering subtitles and also degrade the viewing experience. To further mislead attackers, the video can generate and insert fake subtitle silhouettes into frames that originally contain no subtitles. These artificial subtitles are designed to appear visually similar to real ones but do not contain meaningful text, ensuring they do not impact user comprehension.

{\color{black} To evaluate this defense, we modify the original subtitle (.srt) files by appending padding characters (``\#'') at the end of each subtitle line. This ensures that every frame containing subtitles displays two lines of subtitles, each consisting of 40 characters. This modification preserves the readability of the subtitles while significantly altering their visual appearance. As a result, the subtitles in the recorded videos produce identical silhouette patterns, and \textit{SilhouetteTell} consistently misclassifies these modified silhouettes, incorrectly identifying all randomly selected 25 video clips. Figure~\ref{fig:unify} shows the impact of subtitle padding. Without padding, the original two frames contain one and two lines of subtitles, respectively, resulting in distinguishable subtitle silhouettes in blurry recordings. In contrary, with padding, both blurry frames exhibit identical subtitle silhouettes, effectively confusing \textit{SilhouetteTell}.

\begin{figure}[t]
\centering \includegraphics[width=3in]{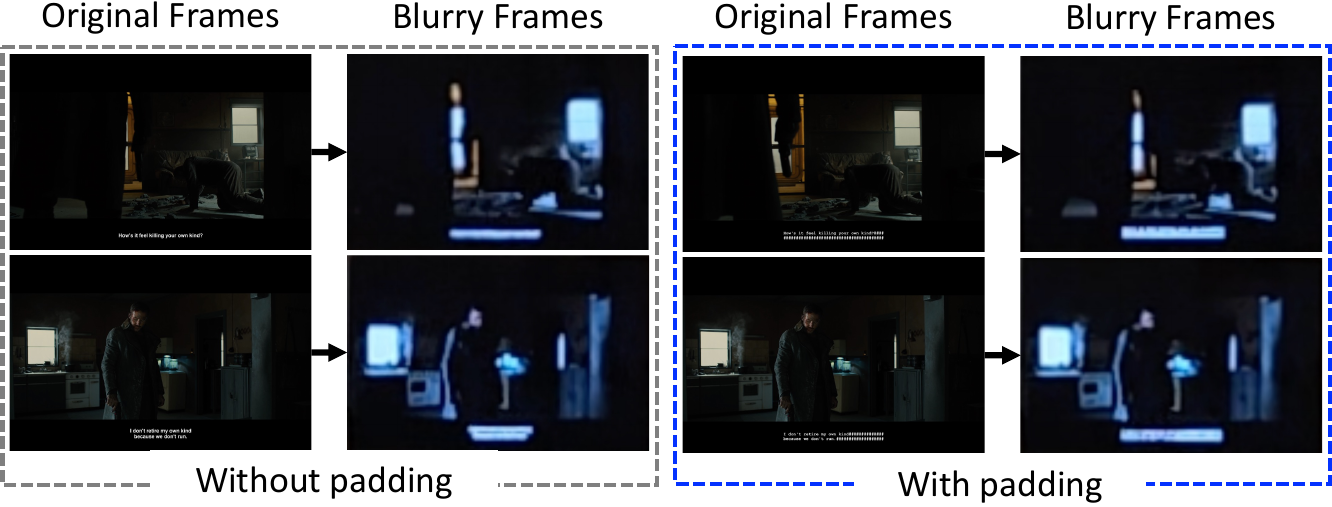} 
\vspace{-0.12in} 
\caption{Illustration of padded subtitles.} 
\label{fig:unify}
\vspace{-0.2in}
\end{figure}

}

{\color{black} \subsubsection{Applying privacy screen protectors} 
Another} defense is to use privacy screen protectors (e.g.,~\cite{Privacy24}) with special filters that allow light to pass through from certain angles (e.g., narrow front angles), which could be further blocked by the victim's body. It would be then difficult for the attacker to find a workable recording angle to video-record clear subtitle silhouettes. However, putting privacy films on screens would reduce the viewing quality for the victim, as the screen's brightness and color may be deteriorated. It also requires the victim to cooperate to help block the angles from which the light is not filtered by the films and can pass through. 

\begin{figure}[t]
\centering \includegraphics[width=2.4in]{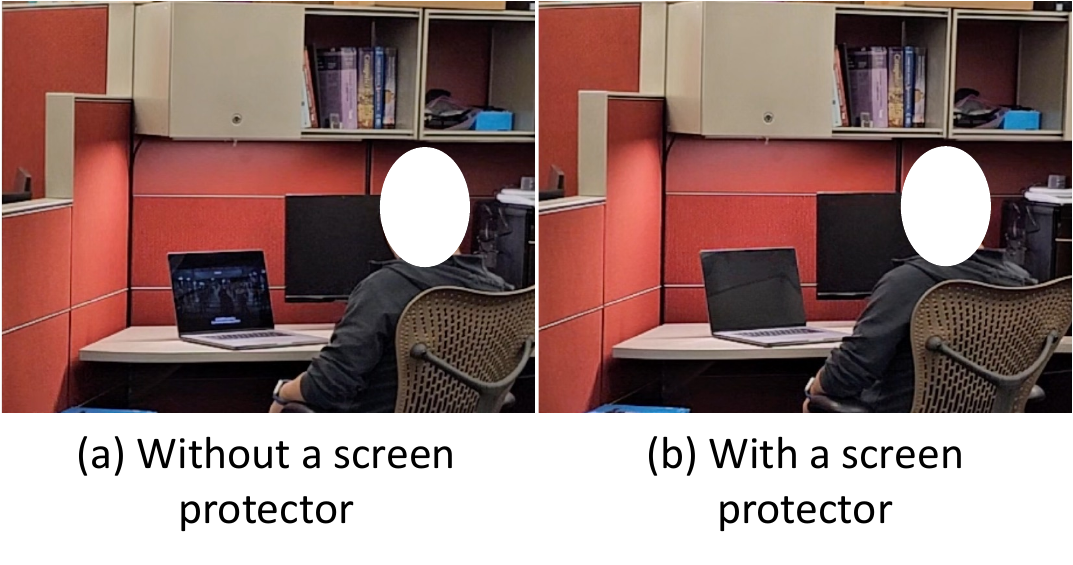}
\vspace{-0.15in} 
\caption{Illustration of the privacy screen protector defense.} 
\label{fig:protector}
\vspace{-0.23in}
\end{figure}

{\color{black}
We evaluate a common privacy screen protector~\cite{peslv_privacy_screen_2025} on a 16-inch MacBook Pro. This protector is designed to reduce screen brightness and limit visibility beyond a horizontal viewing angle of $\pm30\degree$. When the attacker records from outside this range, the captured video appears nearly black, rendering subtitle silhouette extraction infeasible. In this setting, across 25 randomly selected video clips, \textit{SilhouetteTell} again consistently fails to identify the correct videos. Figure~\ref{fig:protector} shows the privacy screen protector's impact. When the protector is in place, the screen becomes effectively invisible from side angles, revealing no subtitle silhouettes.
}

{\color{black} 
\subsection{Potential Extensions}
\label{subsec:extensions}

While \textit{SilhouetteTell} is specifically designed to leverage the spatiotemporal feature of subtitle silhouettes, the underlying methodology can be generalized to other forms of on-screen text. For instance, news tickers commonly appear as horizontally or vertically scrolling text, typically located at the bottom or top of the screen.  By tracking the appearance frequency, shape, and position of these text regions over time, we can construct spatiotemporal fingerprints similar to those used for subtitles. Such fingerprints can then be used for video identification. 
}

\section{Related Work}
\label{sec:related}

\textbf{CV-based Text/image Recognition:} 
By recognizing images or text (e.g., subtitles) in a video, we may infer the video by content matching. Computer Vision (CV) based recognition methods exploit the features from video data and achieve recognizing personnel or objects within images~(e.g., \cite{REN2022102660,MasmoudiA}). OCR and Scene Text Recognition (STR)~\cite{ChenText,WangTowards} are two widely used methods for text recognition. However, all of these methods require clear, detailed, and high-quality recordings to extract text or images accurately, rendering them incapable of handling scenarios when recordings are blurred. On the contrary, our work is the first practical video inference technique that can handle scenarios with blurring recordings when traditional CV-based methods fail. 

{\color{black}
Besides, analyzing reflections from nearby objects has been explored to recover sensitive content displayed on a screen \cite{BackesTempest,raguram2011ispy,xu2013seeing,xu2014watching}. For example, \cite{BackesTempest} is one pioneering study to infer on-screen content from blurred images, while it requires using an astronomical camera that costs over 6,000 USD. In contrast, \textit{SilhouetteTell} works with a low-cost RGB camera. Meanwhile, \cite{raguram2011ispy} and \cite{xu2013seeing} target keystroke inference leveraging key pop-out events and fingertip motion, respectively. These techniques study a different privacy threat from ours. 
Also,~\cite{xu2014watching} infers TV content by leveraging flickering patterns caused by brightness changes due to scene transitions. However, this method is limited to nighttime conditions with no ambient lighting and has only been evaluated on large displays (24/30/50 inches). 
Conversely, \textit{SilhouetteTell} has no such restrictions and remains effective on small screens (e.g., 6.8 inches).
}

\textbf{Traffic Analysis based Video Inference:} 
There have been extensive studies using traffic analysis to infer videos. Different videos may thus result in distinct traffic patterns. In general, traffic-assist video inference attacks can be divided into the following categories according to the source of traffic. (1) \textit{Wired-based: } the attacker may hack the router that the victim connects and then connect to the router through an Ethernet cable for traffic sniffing~\cite{GuWalls,ZhangTraffic}. (2) \textit{WiFi-based: } the attacker may connect to the same WiFi as the victim to capture wireless traffic~\cite{LiDeep,DubinI}. (3) \textit{Cellular-based: } recent studies (e.g.,~\cite{bae2022watching,LakshmananOn}) also show success in achieving video identification by collecting and analyzing LTE traffic. In contrast, our attack does not rely on traffic for video inference and works regardless of whether the victim watches videos online or offline. 

\section{Conclusion}
\label{sec:conclusion}

We propose \textit{SilhouetteTell}, a novel and practical video inference technique, with the following advantages over previous methods. (1) Non-invasive: there is no need to pre-infect the victim's device with malware. (2) Traffic-independent: it works for both online and offline video streaming. (3) No user-specific training is needed: no labeled data is needed from the victim and it can handle cases when traditional text/image recognition algorithms fail. (4) It requires no close proximity: it can be launched several meters, even tens of meters, away from the target.  \textit{SilhouetteTell} is the first to identify and build the spatiotemporal correlation between the captured blurry video frames and embedded subtitles. Our extensive evaluation on top of off-the-shelf smartphones verifies that \textit{SilhouetteTell} can achieve high accuracy in identifying video titles and clips that the victim is watching from a distance of up to 40 meters. 

\begin{acks}

The authors would like to thank all anonymous reviewers for their insightful comments. This work was supported in part by the National Science Foundation under Grants No. 2155181 and No. 2424439.
\end{acks}

\bibliographystyle{ACM-Reference-Format}
\bibliography{main}

\appendix

\begin{figure*}[t]
\centering \includegraphics[width=5.5in]{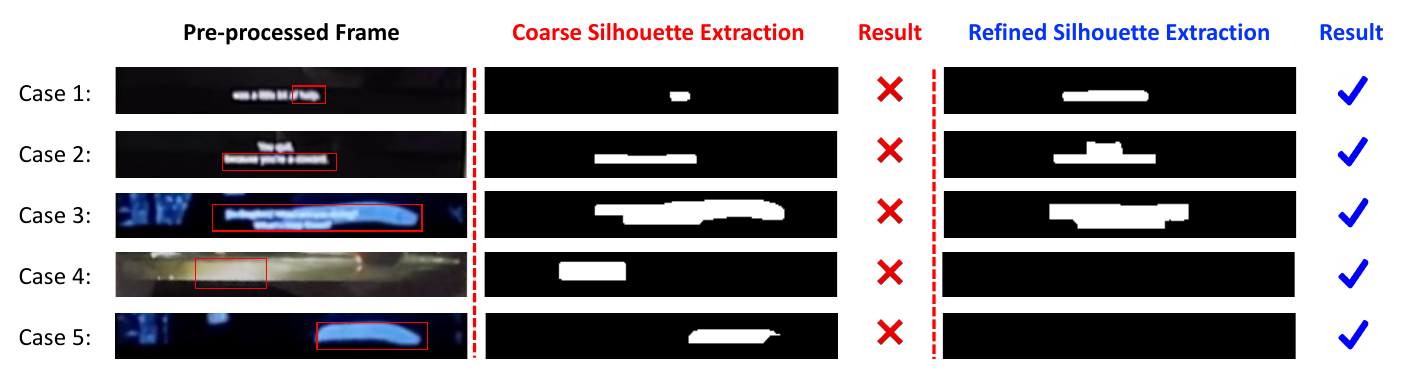}
\vspace{-0.12in} 
\caption{Comparison of extraction results between coarse and refined silhouette extraction.}
\label{fig:cmp_two}
%\vspace{-0.05in}
\end{figure*}

\section{Subtitle Data Crawling}
\label{appendix:crawling}
%\textit{Subtitle Data Crawling: }
We can directly copy the URL of a YouTube video to web applications such as DownSub~\cite{DownSub}, and then download its subtitles. We can also write a script to automatically get the subtitles for a given YouTube video using a Python API  (i.e., youtube-transcript-api~\cite{api24}). Besides, the following steps can be performed to obtain subtitle files from Amazon Prime Video and Netflix: (1) open DevTools (i.e., a set of web developer tools) from Chrome menus~\cite{DevTools}; (2) in the Network tab, filter ``.ttml2" (for Amazon Prime Video) or a string ``?o=" (for Netflix) when playing a video online; (3) capture subtitle files from the returned results and further convert them into .srt files via a subtitle converter tool (e.g.,~\cite{SubConverter}).

\section{Silhouette Extraction Comparison}
\label{appendix:com}

%as shown in 
Figure~\ref{fig:cmp_two} compares coarse and refined silhouette extraction methods in dealing with five different cases, where coarse silhouette extraction fails while refined silhouette extraction manages to generate correct recognition results. In cases 1-3, coarse silhouette extraction makes errors in recognizing silhouette width, height, and both width and height, respectively; in cases 4-5, the input frames have no subtitles, while coarse silhouette extraction incorrectly recognizes interference shapes as subtitle silhouettes.

\section{Impact of Recording Resolution} 
\label{subsubsec:resolution} 

Usually, external subtitles (e.g., SRT files) of a video are independent and separate from the video, and their resolution (i.e., quality) will not change with the resolution of the video being played. Therefore, if the video resolution changes, the silhouettes of the subtitles of this video may not change accordingly. On the contrary, the adversary may utilize different resolutions to do recording, capturing silhouettes that may have inconsistent clarity. We consider three typical recording resolutions for the recording device (Samsung Galaxy Z Fold4): 720p (1280$\times$720), 1080p (1920$\times$1080), and 4k (3840$\times$2160). For each resolution, we randomly record 100 video clips. 

Table~\ref{table:res} presents the top-$k$ accuracy for inferring video titles and clips under different recording resolutions. We can see our attack can consistently obtain pretty high (above 90\%) inference performance. Particularly, the top-10, top-20, and top-50 accuracy values are always 1 regardless of the recording resolution. Also, with the resolution decreasing, the top-1 accuracy slightly decreases, and top-5 accuracy almost maintains a high value (i.e., 99.5\%). 

\begin{table}[t] 
 \caption{Average top-$k$ accuracy for video titles and clips (referred to as ``T" and ``C") vs. recording resolution.}
 \vspace{-0.15in}
 \centering
 \small
\begin{tabular}{| *{16}{c|}  }
 \hline
 \!\multirow{2}{*}{Resolution}\!	& \multicolumn{2}{c|}{Top-1} & \multicolumn{2}{c|}{Top-5} & \multicolumn{2}{c|}{Top-10} & \multicolumn{2}{c|}{Top-20}  & \multicolumn{2}{c|}{Top-50}  \\
\cline{2-11} 
		&\! \!T\!\! 	&\! \!C\!  	& \!\!T\!\! 	&\! \!C\! 	& \!\!T\!\!  	&\!\!C\!\!  	&\!\!T\!\! 	&\!\!C\!\!		& \!\!T\!\!		&\!\!C\!\!\\ 
 \hline 
4k 		& \!\!0.953\!\! 	&  \!\!0.953\! 	& \!\!0.995\!\!  	&\! \!0.995\!		& \!1.0\!  	&\!1.0\! 	& \!1.0\!	&\!1.0\!		&\! 1.0\!		&\!1.0\! \\  
  \hline 
 1080p 	& \!\!0.937\!\! 	& \!\!0.937\! 	& \!\!0.995\!\!  	&\! \!0.995\! 	& \!1.0\!  	&\!1.0\!	& \!1.0\!	&\!1.0\!		& \!1.0\!		&\!1.0\! \\ 
\hline
720p 	&\! \!0.903\!\! 	&\! \!0.901\!  	& \!\!0.995\!\! 	&\! \!0.994\! 	& \!1.0\!  	&\!1.0\!  	&\!1.0\! 	&\!1.0\!		& \!1.0\!		&\!1.0\!\\ 
 \hline
\end{tabular}
\label{table:res}
% \vspace{-0.1in}
\end{table}

\begin{figure}[t]
\centering \includegraphics[width=2.9in]{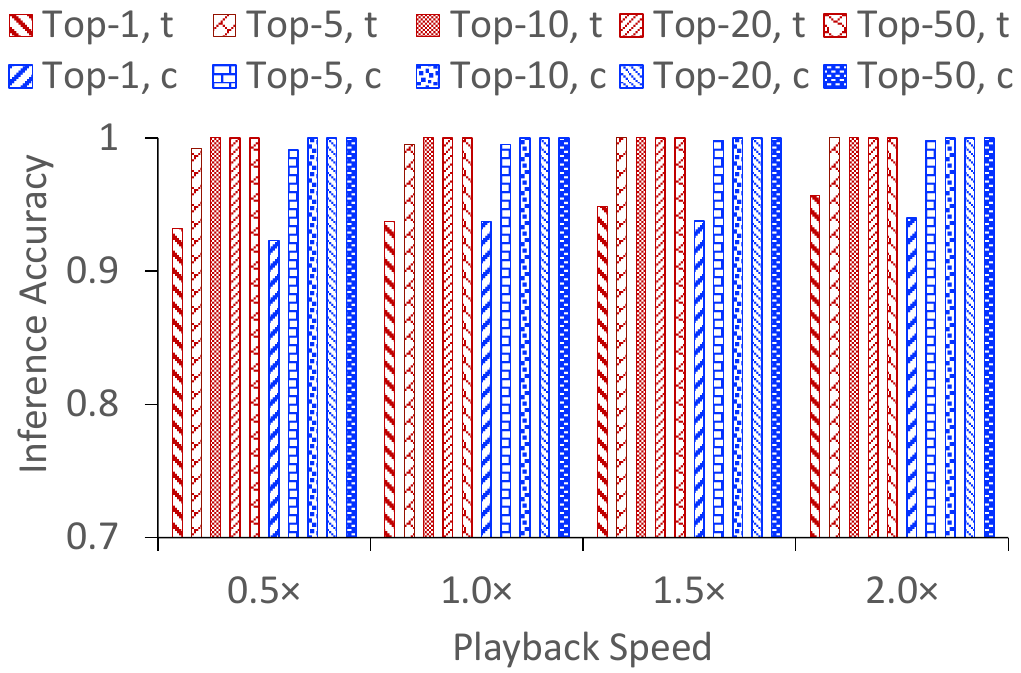}
\vspace{-0.1in} 
\caption{Impact of playback speed.}
\label{fig:play_speed}
\vspace{-0.05in}
\end{figure}

\section{Impact of Playback Speed} 
\label{subsubsec:speed} 

We test four different playback speeds: 0.5$\times$, 1.0$\times$, 1.5$\times$, and 2.0$\times$. For each playback speed, we randomly record 100 2-minute video clips while keeping other factors (e.g., recording distance, angle, and resolution) consistent. Figures~\ref{fig:play_speed} present the resultant inference performance. We observe that for both video titles and clips, the top-1 and top-5 accuracy values slightly increase as the playback speed increases, while the top-10, top-20, and top-50 accuracy values remain at 100\%. This improvement mainly results from an increase in the length of the obtained subtitle vector at higher playback speeds, which reduces the number of candidates. Particularly, the top-1 accuracy for video titles ranges from 93.3\% to 95.7\% regardless of the playback speed. 

\begin{figure}[t]
\centering \includegraphics[width=2.3in]{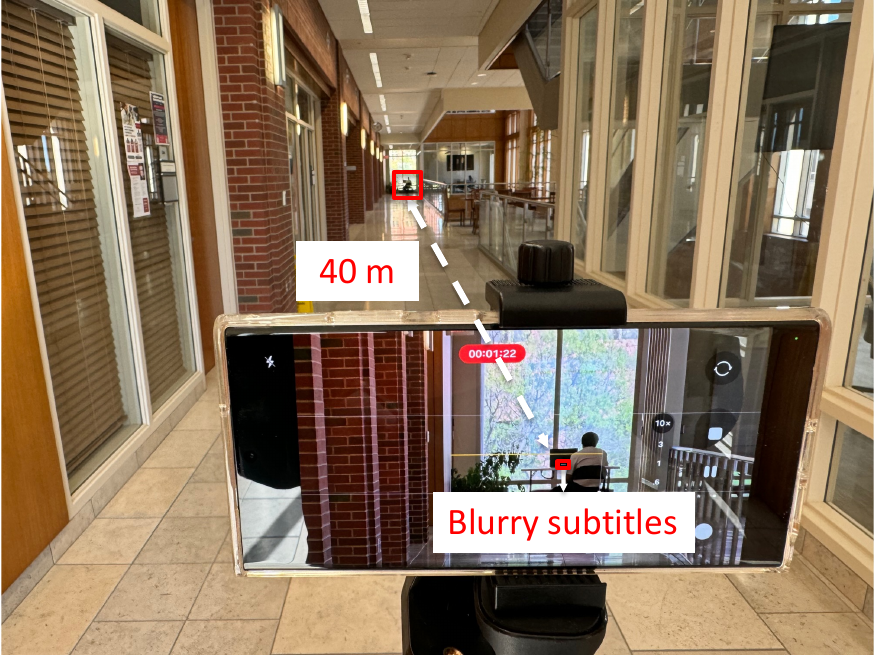} 
\vspace{-0.1in} 
\caption{Recording the target across the 40 m corridor.} 
\label{fig:long_distance}
%\vspace{-0.05in}
\end{figure}

\section{Long-distance Testing Environment} 
\label{appendix:long-distance}

Figure~\ref{fig:long_distance} shows our long-distance testing environment. 

\section{Scalability Analysis} 
\label{appendix:scalability_analysis}

\begin{figure}[t]
\vspace{-0.05in} 
\centering \includegraphics[width=3.5in]{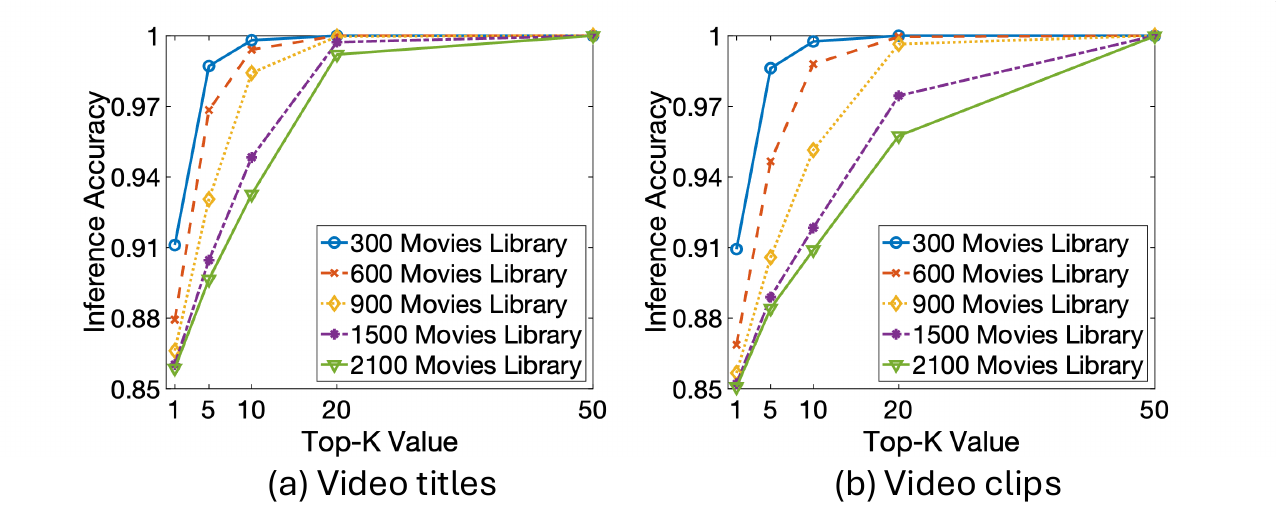} 
\vspace{-0.3in} 
\caption{Average top-\textit{k} accuracy vs. library size.} 
\label{fig:scala}
\vspace{-0.05in}
\end{figure}

\textbf{Scalability Analysis} 
In existing video inference studies~\cite{GuWalls,ReedIdentifying,ZhangTraffic,bae2022watching}, the number of videos used typically ranges from 100 to 300. Due to the widespread availability of subtitle files, \textit{SilhouetteTell} can easily expand its video database for matching. We examine the effects of video libraries of five different sizes: 300, 600, 900, 1,500, and 2,100. Accordingly, we randomly select 100, 200, 300, 500, and 700 videos from each video provider. The video durations range from 24 to 242 minutes. Figure~\ref{fig:scala} (a) and (b) illustrate the corresponding average video titles and clips inference accuracy. We observe that increasing the library size does not significantly impact inference accuracy. Particularly,
the top-1 and top-20 accuracy values remain above 85\% and 96\%, respectively, across all video library sizes.

\end{document}